\pdfoutput=1
\AddToHook{cmd/maketitle/after}{%
  \begingroup
  \footnotetext{\raggedright Findings of the Association for Computational \mbox{Linguistics: EACL 2027.}}%
  \endgroup
}

\documentclass[11pt]{article}

\usepackage[final]{acl}

\usepackage{times}
\usepackage{latexsym}
\usepackage{caption}

\usepackage[T1]{fontenc}
\usepackage[utf8]{inputenc}
\usepackage{microtype}

\usepackage{graphicx}

\usepackage{hyperref}
\usepackage{url}
\usepackage[dvipsnames]{xcolor}
\usepackage{multirow}

\usepackage{subcaption}
\usepackage{float}
\floatstyle{ruled}
\newfloat{algorithm}{t}{loa}
\floatname{algorithm}{Algorithm}
\usepackage{algpseudocode}
\usepackage{algorithmicx}
\usepackage{amsmath}
\usepackage{amssymb}
\usepackage{amsthm}
\usepackage{amsfonts}
\usepackage{booktabs}
\usepackage{cleveref}
\theoremstyle{plain}
\newtheorem{theorem}{Theorem}
\newtheorem{lemma}{Lemma}

\theoremstyle{definition}
\newtheorem{definition}{Definition}

\definecolor{linkblue}{HTML}{1E6BD6}

\definecolor{revisionblue}{HTML}{145AB6}
\newcommand{\revision}[1]{#1}

\algrenewcommand\algorithmicrequire{\textbf{Input:}}
\algrenewcommand\algorithmicensure {\textbf{Output:}}
\definecolor{jlcolor}{HTML}{16825D}

\newcommand{\lz}[1]{}

\crefname{section}{Section}{Sections}
\Crefname{section}{Section}{Sections}
\crefname{figure}{Figure}{Figures}
\Crefname{figure}{Figure}{Figures}
\crefname{table}{Table}{Tables}
\Crefname{table}{Table}{Tables}
\crefname{equation}{Eq.}{Eqs.}
\Crefname{equation}{Eq.}{Eqs.}
\crefname{theorem}{Theorem}{Theorems}
\Crefname{theorem}{Theorem}{Theorems}
\crefname{lemma}{Lemma}{Lemmas}
\Crefname{lemma}{Lemma}{Lemmas}
\crefname{algorithm}{Algorithm}{Algorithms}
\Crefname{algorithm}{Algorithm}{Algorithms}
\crefname{appendix}{Appendix}{Appendices}
\Crefname{appendix}{Appendix}{Appendices}

\title{SPECTRA: Revealing the Full Spectrum of \\ LLM-Internal Preference Distributions}

\author{
    \textbf{Luyang Zhang\textsuperscript{\textdagger}} \thanks{Work conducted while interning at TikTok. Correspondence: \texttt{luyangz@andrew.cmu.edu}},
    \textbf{Jialu Wang\textsuperscript{\textdaggerdbl}},
    \textbf{Shichao Zhu\textsuperscript{\textdaggerdbl}},
    \textbf{Beibei Li\textsuperscript{\textdagger}} \\
    \textbf{Zhongcun Wang\textsuperscript{\textdaggerdbl}},
    \textbf{Guangmou Pan\textsuperscript{\textdaggerdbl}},
    \textbf{Yang Song\textsuperscript{\textdaggerdbl}}
    \\
    \textsuperscript{\textdagger} Carnegie Mellon University
    \\
    \textsuperscript{\textdaggerdbl} TikTok Inc.
}

\begin{document}
\maketitle

\begin{abstract}
\revision{Large language models (LLMs) often represent user preferences by
generating ranked item lists. These lists can overrepresent frequent
categories and omit long-tail interests. We instead treat preference
prediction as an LLM classification task, where a fine-tuned LLM assigns
a probability to each category. Given a user's interaction
history, we estimate the share of future interactions in each category.
We propose SPECTRA (Softmax Probing for Extracted Category-level
Token Readouts and Analysis), which reads a distribution over
interpretable preference categories from a fine-tuned LLM's token scores.}
We evaluate SPECTRA on MovieLens, Yelp, and a large-scale short-video
platform. SPECTRA delivers (i) distributional alignment, reducing
Jensen-Shannon divergence to the empirical preference distribution by
38--44\% across public datasets; (ii) long-tail recovery with
cross-user fairness, raising top-3 category exposure entropy by
$23\%$ on MovieLens and producing a larger gain on tail-preference
users than on head-preference users; and (iii) downstream application
value, with a 41--46\% category-NDCG boost on MovieLens and Yelp,
and an increase in long-tail NDCG@$20$ from $0.024$ to $0.197$
against a production ranker on logged short-video data.
\end{abstract}

\section{Introduction}
\label{sec:intro}

Understanding user preferences in an explicit and interpretable way is
a central challenge in personalized AI~\citep{doshi2017towards,
amershi2019guidelines,rudin2019stop,lipton2018mythos}. Large Language
Models (LLMs) are a promising tool for this task, given their ability
to integrate text content (reviews, descriptions, category labels) with
user behavior data. However, extracting an explicit, structured
preference distribution from an LLM requires treating it not just as a
generator of items, but as a probabilistic model over user
preferences~\citep{wang2023large,li2023transformers}.
Such distributions, defined over semantically interpretable preference categories,
can then be analyzed, audited, and reused across downstream
applications.

The prevailing approach of using LLMs to directly generate ranked item
lists~\citep{ngo2024recgpt,deng2025onerec,wang2025user}
leaves two critical gaps. First, the model's preference state stays
hidden. Direct generation (DG) produces a single ranked list, not a
structured distribution, leaving the user's underlying preferences
uninterpretable. Second, the output step itself loses information,
keeping only the top few
candidates~\citep{holtzman2019curious,stahlberg2019nmt} and discarding
low-probability categories, systematically suppressing minority and
long-tail preferences. We theoretically show that once a decoded list
omits a category, any downstream distribution formed only from that
list remains structurally unable to assign it mass. Under repeated deployment, this loss compounds at
scale~\citep{bakshy2015exposure}.

\revision{To close these gaps, we treat the category of a user's future
interaction as a classification target and use a fine-tuned LLM as
the classifier. Decision models such as Jev~\citep{almeida2026jev},
released one year after our initial submission in September 2025,
share this design and return probabilities over predefined answers in
place of generated text. SPECTRA (Softmax Probing for Extracted Category-level
Token Readouts and Analysis) predicts a distribution over interpretable
categories such as movie genres, video tags, or restaurant categories.
We score each category with the LLM, either from the
difference between affirmative and negative logits in a yes/no probe
or from the logit of its answer token in a multiple-choice prompt.
A softmax over the category scores yields the preference distribution.}
When the category space is large, we develop a hierarchical pipeline that
first probes coarse-grained categories and then refines into finer
subcategories within the selected categories. We prove that SPECTRA recovers the true top-$k$ categories when
its per-category prediction error is less than half the gap between
the $k$th and $(k+1)$th categories. We also bound the probability
that frequency-biased DG misses a true top-$k$ category.

We empirically evaluate SPECTRA on distributional alignment, long-tail
exposure, and downstream ranking. First, SPECTRA assigns
probability to long-tail categories omitted by DG and raises
top-3 category exposure entropy by $23\%$ on MovieLens. Second, SPECTRA's gain in matching the true distribution is 13--28\%
larger for users with niche preferences. Third, SPECTRA performs even better when the category space is large.
Its NDCG advantage over DG widens by roughly $4\times$ on Yelp's large
category space ($1{,}000{+}$ categories) compared to MovieLens.
\revision{Finally, on a shared short-video test set, SPECTRA
predicts whether users view each category more accurately than
Kev-4B~\citep{palmer2026kev4b} and the alternative readouts
evaluated in \cref{sec:exp_readout_comparison}.}

Our contributions are threefold:
\begin{itemize}
\item \textbf{Method.} We propose a framework that probes an LLM's logits to
model user preferences over the space of categories, rather than
decoding a ranked list of items.

\item \textbf{Empirics and theory.} We empirically demonstrate that
probing yields more faithful preference distributions, more balanced
coverage of long-tail preferences, and stronger downstream value than
the evaluated DG baselines, supported by regime-conditional guarantees
that SPECTRA recovers true top categories while sparse list decoding
can miss the tail.

\item \textbf{Impact.} A theory-grounded approach toward more
interpretable and balanced user preference modeling with LLMs,
delivering named-category distributions that preserve minority
preferences rather than collapsing them at decoding.
\end{itemize}

The SPECTRA repository contains the probing code and a runnable example
(\url{https://github.com/LZhang12345/spectra}).

\section{Related Work}
\label{sec:related}

\textbf{Probing LM Distributions for Latent Beliefs.}
Belief extraction from token likelihoods has been studied for
factual knowledge probing and prompt-based classification
\citep{petroni2019language,schick2020s,raffel2020exploring}.
Direct decoding suffers from search artifacts
\citep{holtzman2019curious,stahlberg2019nmt}, motivating inference
that avoids decoding heuristics.
\revision{Statistics from sampled outputs on labeled validation data can
also predict which model configurations benefit most from best-of-$N$
decoding~\citep{zhang2026predictinginferencetimescalinggains}.}
Distributional
objectives in alignment and uncertainty modeling
\citep{siththaranjan2023distributional,yao2024no,melnyk2024distributional}
further motivate predicting an explicit preference distribution.
\revision{Jev~\citep{almeida2026jev} returns probabilities over predefined
answers to structured questions.}
Scaling to large category spaces commonly uses coarse-to-fine
hierarchical inference, as in extreme multi-label classification
\citep{zhou2024quest}.
SPECTRA differs in two ways. Unlike
\citet{zhou2024quest}, which probes a frozen LM's existing factual
taxonomies via token-likelihood ranking, SPECTRA first aligns the LM
through CPT$+$SFT on user interaction data so that the same probing
mechanism reads a per-user preference signal rather than encyclopedic
priors. Unlike \citet{siththaranjan2023distributional}, which models
distributional preferences across annotators in the reward-modeling
setting, SPECTRA recovers a distribution \emph{per individual user}
over a category vocabulary from the LM's softmax.

\textbf{LLMs for Preference Modeling.}
LLMs have been applied to preference modeling through prompt-based
unification (P5; \citep{geng2022recommendation}), text-aware user
representations (U-BERT; \citep{qiu2021u}), and generative
recommendation or exploration
\citep{ngo2024recgpt,deng2025onerec,wang2024llms,wang2025user,yi2025recgpt}.
Recent extensions ground preferences in review text
\citep{kim2025exp3rt}, contextualize user embeddings inside the LLM
\citep{ning2025userllm}, or enhance generative retrieval
\citep{paischer2025preference}, while reasoning-style pipelines
produce item-level rationales \citep{yue2025cot4rec,wang2024llmrg}.
Unlike P5 and U-BERT, which compress user state into prompt
templates or learned embeddings, and unlike CoT4Rec and LLMRG, which
produce item-level rationales or rankings, SPECTRA reads an explicit
\revision{category-level distribution from the aligned LM's token scores
without decoding a ranked list.}

\textbf{User Preference Representation.}
Prior work models user preferences from interaction histories using
\emph{multi-preference} representations
\citep{shi2023everyone,cen2020controllable,zhou2018deep} and
\emph{sequential} next-item predictors
\citep{rendle2010factorizing,hidasi2015session,kang2018self,sun2019bert4rec}.
These methods typically encode user state implicitly as embeddings or
item-level scores, which limits transparency for analysis and
auditing
\citep{doshi2017towards,lipton2018mythos,rudin2019stop}.
Unlike multi-preference encoders, which represent user state as a
fixed number of latent vectors, and unlike sequential next-item
predictors such as SASRec \citep{kang2018self} and GRU4Rec
\citep{hidasi2015session}, which output items rather than category
distributions, SPECTRA produces a single category-level distribution
per user that is directly inspectable, comparable across cohorts and
time, and downstream-composable with classical recommenders as a
category prior.
\revision{In sponsored listings, \citet{lu2025powerlinearprogrammingsponsored}
formulate downstream ranking as a constrained optimization problem that
balances revenue and relevance.}

\section{Method}
\label{sec:method}

\begin{figure*}[t]
\centering
\includegraphics[width=\textwidth]{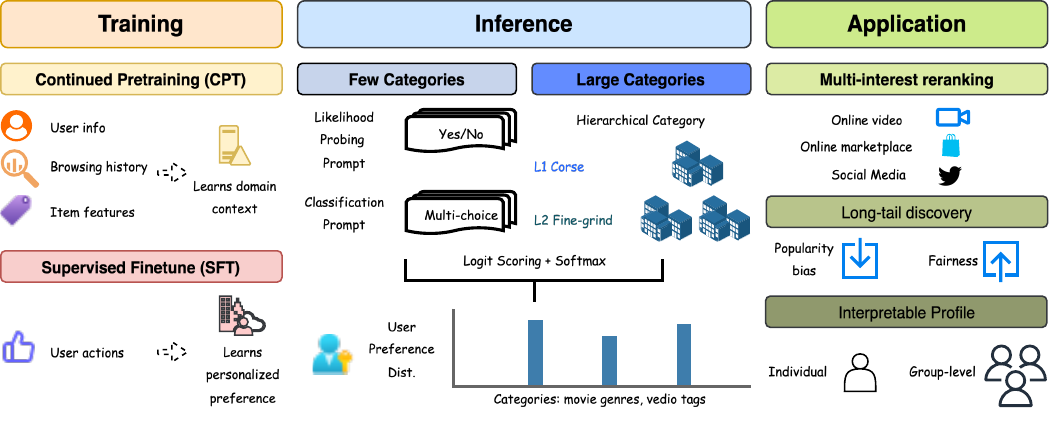}
\caption{Overview of SPECTRA. \textbf{Training (left).} Domain
continued pretraining (CPT) followed by LoRA supervised fine-tuning
(SFT). \textit{Inference (middle).} SPECTRA reads a per-category
score vector from the model's softmax via likelihood probing or
generative classification, then normalizes to a category-level
preference distribution. A hierarchical variant handles large
taxonomies. \textit{Application (right).} The preference
distribution supports reranking, long-tail discovery, and
interpretable user profiles.}
\label{fig:framework}
\end{figure*}

We first formalize the problem of modeling a user's preference
distribution over item categories (\cref{sec:method:setup}). We then describe SPECTRA, a method that
extracts a category-level preference distribution from a fine-tuned
LLM (\cref{sec:method:primitive}). Finally, we theoretically compare
SPECTRA and DG in terms of their distance to the true
preference distribution (\cref{sec:method:theory}).
\Cref{fig:framework} gives a visual overview of the full pipeline:
domain training (CPT $+$ LoRA SFT), inference (likelihood-based
probing or generative classification, with a hierarchical variant for
large category spaces), and downstream applications (recommendation,
long-tail discovery, interpretable user profiles).

\subsection{Setup and notation}
\label{sec:method:setup}

Fix a set of $K$ semantic categories
$C = \{c_1, \dots, c_K\}$ (e.g., movie genres, restaurant categories),
shared across users and time. For each user, let $X_{1:t}$ denote the
user's interaction history up to a reference time $t$. Our goal is to
predict the user's preference distribution over $C$ for a future
window $[t{+}1, T]$, i.e., the fraction of the user's future
interactions that fall into each category. We define two
probability distributions over $C$.
\begin{itemize}
\item $\theta_*$ is the empirical ground-truth distribution,
obtained by counting the user's interactions in each category over
$[t{+}1, T]$.
\item $\theta_S$ is the distribution produced by SPECTRA
(\cref{sec:method:primitive}), defined over all $K$ categories.
\end{itemize}

\subsection{The SPECTRA method}
\label{sec:method:primitive}

SPECTRA consists of an \textit{alignment} stage that fine-tunes the
LLM to encode preference distributions over $C$, and an
\textit{inference} stage that extracts a per-category distribution
from the aligned LLM.

\textbf{Alignment (CPT + LoRA SFT).}
\textit{Continued pretraining (CPT)} on textualized user and item
metadata and interaction traces teaches the LLM the domain context
(user behaviors, item types, domain vocabulary). \textit{LoRA SFT}
then trains the LLM to predict the user's long-term preference
distribution over $C$ (computed empirically from interactions in
$[t{+}1, T]$) rather than a single next-item label. The SFT
objective is standard next-token cross-entropy on interaction
sequences whose targets are category-name tokens. At inference,
the probes below score affirmative and negative tokens or
multiple-choice answer tokens.

\textbf{Inference.}
Given context $X_{1:t}$ and a textual label for each category $c_j$,
SPECTRA constructs a logit score vector $S \in \mathbb{R}^K$ and outputs
$\theta_S := \operatorname{softmax}(S / \tau)$. We compute $S$ in two
ways.

\textit{(1) Likelihood-based probing.}
For each category $c_j$, we pose a yes/no probe under a template
prompt $\pi_{\text{probe}}(X_{1:t}, c_j)$
(\cref{sec:appendix:robustness:prompt}) and read the LLM's
next-token logit vector $\mathbf{z}_j \in \mathbb{R}^{|V|}$ over its
token vocabulary $V$. Using small affirmative and negative token
sets $V_+$ (e.g., \texttt{Yes}, \texttt{yes}) and $V_-$ (e.g.,
\texttt{No}, \texttt{no}),
\begin{equation}
S_j \;:=\; \frac{1}{|V_+|}\!\sum_{v \in V_+}\!\mathbf{z}_j[v]
        \;-\; \frac{1}{|V_-|}\!\sum_{v \in V_-}\!\mathbf{z}_j[v]
\label{eq:logit-score}
\end{equation}
(\cref{alg:spectra-likelihood} in \cref{sec:appendix:algorithm}).

\textbf{(2) Generative classification.}
We construct a single multi-choice prompt
$\pi_{\text{gen}}(X_{1:t}, C)$
(\cref{sec:appendix:robustness:prompt}) that shares the same
context-encoding template as $\pi_{\text{probe}}$ but asks a
multi-choice question over $C$ rather than a yes/no question per
category. The prompt maps each $c_j$ to an answer token $v_j$
(e.g., \texttt{A}, \texttt{B}, \dots). The LLM produces a next-token
logit vector $\mathbf{z} \in \mathbb{R}^{|V|}$, and we set
$S_j := \mathbf{z}[v_j]$
(\cref{alg:spectra-generative} in \cref{sec:appendix:algorithm}).
SPECTRA is robust to prompt form
(\cref{sec:appendix:robustness:prompt}).

\textbf{Scaling via hierarchical probing.}
Likelihood-based probing costs $O(K)$ LLM forward passes per user,
which becomes expensive for large category sets. A hierarchical
variant probes the category set in multiple layers, coarse to fine. When
$C$ carries a two-level taxonomy
$C = \bigsqcup_{a \in C_{L1}} C_{L2}(a)$, we factorize the readout
as
\begin{equation}
\theta_S(c) \;=\; \theta_S\!\bigl(L_1(c)\bigr)\,\cdot\,
                   \theta_S\!\bigl(c \,\mid\, L_1(c)\bigr),
\label{eq:hier-factorization}
\end{equation}
and proceed top-down. \textit{L1 probing} estimates a distribution
over L1 categories. \textit{L2 exploration} then probes L2
categories only within a subset of L1 branches $\mathcal{G}$
(high-mass branches for relevance, long-tail branches for
exploration, depending on the application). This reduces cost to
$O\!\bigl(|C_{L1}| + \sum_{g \in \mathcal{G}} |C_{L2}(g)|\bigr)$
when only a subset of branches is probed. If the taxonomy is a
partition, \cref{eq:hier-factorization} gives a full distribution
when all branches are expanded, as in the reported Yelp evaluation
(\cref{sec:appendix:e1_e5:yelp_hierarchy}).

\subsection{Theoretical comparison}
\label{sec:method:theory}

We theoretically compare SPECTRA and DG on a general
family of ranking-quality metrics.

\textbf{Notation.}
Recall $\theta_*$ from \cref{sec:method:setup} and $\theta_S$ from
\cref{sec:method:primitive}. Let $L \subseteq C$ denote DG's decoded
top-$k$ list ($|L| = k < K$). Define truth's top-$k$ as $G^* :=
\mathrm{top}_k(\theta_*)$, SPECTRA's induced top-$k$ as $L_S :=
\mathrm{top}_k(\theta_S)$, and the \emph{ranking gap} at position $k$
as
\begin{equation}
\Delta_k \;:=\; \theta_*^{(k)} - \theta_*^{(k+1)},
\label{eq:ranking-gap}
\end{equation}
the separation between the $k$-th and $(k{+}1)$-th largest values of
$\theta_*$. As a canonical ranking-quality metric we use the mass captured by a
size-$k$ list,
\begin{equation}
M(\pi, \theta_*) \;:=\; \sum_{c \in \pi} \theta_*(c),
\label{eq:mass-captured}
\end{equation}
one member of the Schur-isotone family of ranking metrics, which
also includes NDCG@$k$ and DCG@$k$
(\cref{sec:appendix:ranking-advantage:schur}).

Under these definitions, we state the main theorem on SPECTRA's
ranking advantage over DG.

\begin{theorem}[SPECTRA ranking advantage]
\label{thm:ranking-advantage}
For a fixed user, suppose (i) SPECTRA's readout has bounded
pointwise error, $\|\theta_S - \theta_*\|_\infty \leq \varepsilon$ for
some $\varepsilon \geq 0$; (ii) the ranking gap is lower-bounded by
twice this error, $\Delta_k > 2\varepsilon$; and (iii) DG does not
reproduce the true top-$k$, $L \neq G^*$. Then $L_S = G^*$ and
\begin{equation}
M(L_S, \theta_*) \;>\; M(L, \theta_*).
\label{eq:ranking-advantage}
\end{equation}
See \cref{sec:appendix:ranking-advantage}.
\end{theorem}

\textbf{Interpretation of the theorem.}
When SPECTRA's per-category readout is closer to the truth than half
the ranking gap, and DG's decoded list misses any true-top-$k$
category, SPECTRA's top-$k$ recovers truth's top-$k$ exactly and
strictly outperforms DG on the mass-captured metric. Geometrically,
$\theta_S$ lies within an $\varepsilon$-ball of $\theta_*$ on the
category simplex, whereas $\theta_{\mathrm{DG}}$ is confined to the
lower-dimensional face supported on $L$ and so cannot place any mass
on categories outside $L$.

\Cref{thm:ranking-advantage} takes the DG-miss condition (iii) as a
hypothesis. We next give a complementary lower bound that derives
condition (iii) from a structural property of the decoding
distribution, so that the two theorems together form a two-sided
separation result.

\begin{definition}[Frequency-biased decoding]
\label{def:freq-bias}
The decoding distribution $\pi_{\mathrm{DG}}$ over $C$ has
frequency bias with coefficient $\lambda \geq 0$ if for all
$c \in C$,
\begin{equation}
\log \pi_{\mathrm{DG}}(c)
\;=\; z(c) \,+\, \lambda \log f(c) \,+\, C_0,
\label{eq:freq-bias}
\end{equation}
where $z(c)$ is the LLM's preference logit, $f(c)$ is the empirical
pretraining frequency of $c$'s category-name token (with standard
smoothing so $f(c)>0$), and $C_0$ normalizes the distribution.
Coefficient $\lambda > 0$ captures the empirically documented
head-frequency amplification in autoregressive LM decoding
\citep{zhao2021calibrate,holtzman2019curious}.
\end{definition}

\begin{theorem}[DG miss-probability lower bound]
\label{thm:dg-miss}
Suppose $\pi_{\mathrm{DG}}$ satisfies \cref{def:freq-bias} with
coefficient $\lambda > 0$, and the true top-$k$ list $G^*$ contains
a category $c^*$ whose pretraining frequency satisfies
$f(c^*) / \max_{c \in C} f(c) \leq \rho^{-1}$ for some $\rho > 1$.
Assume the LLM's preference logits are bounded,
$|z(c^*) - z(c_{\mathrm{head}})| \leq B$, where $c_{\mathrm{head}}$
is the argmax-frequency category. Then for any category-selection
step whose distribution satisfies \cref{def:freq-bias},
\begin{equation}
\frac{\pi_{\mathrm{DG}}(c^*)}{\pi_{\mathrm{DG}}(c_{\mathrm{head}})}
\;\leq\; e^B \,\rho^{-\lambda},
\label{eq:per-step-bound}
\end{equation}
If DG forms a length-$k$ list through category-selection steps that
obey the same bound conditional on earlier selections, a union bound
gives that the probability DG includes $c^*$ in its top-$k$ list is at
most $k \, e^B \rho^{-\lambda}$. Since $c^* \in G^*$, any list $L$
that omits $c^*$ satisfies $L \neq G^*$, hence
\begin{equation}
\Pr\!\bigl[\,L \neq G^*\bigr]
\;\geq\; 1 - k \, e^B \,\rho^{-\lambda}.
\label{eq:dg-miss}
\end{equation}
\end{theorem}

\Cref{thm:dg-miss} is a worst-case structural bound that
characterizes when DG is provably likely to miss tail categories;
we do not fit $\lambda, \rho, B$ as predictive parameters but use
them to delineate the failure regime that the empirical decoding
literature documents \citep{holtzman2019curious,zhao2021calibrate}.

\textbf{Combined separation.}
\Cref{thm:ranking-advantage} and \cref{thm:dg-miss} together
characterize a regime in which SPECTRA strictly improves over DG.
Whenever SPECTRA's calibration satisfies (i) and (ii) of
\cref{thm:ranking-advantage}, and the true top-$k$ contains at least
one category that is rare in pretraining (frequency ratio
$\rho^{-1}$), SPECTRA recovers $G^*$ exactly while DG fails to
recover it with probability at least $1 - k e^B \rho^{-\lambda}$.
The DG-miss condition (iii) is therefore a consequence of frequency
bias, not an assumption. Long-tail categories with low pretraining
frequency are precisely the regime where this bound is tight, which
matches the empirical pattern in \cref{sec:experiments}.

\textbf{What the theory licenses.}
The separation is intentionally regime-conditional: it isolates a
support-mismatch failure mode in which sparse decoded lists cannot
represent categories outside their support, while a calibrated softmax
readout can assign mass over the whole category vocabulary.

\textbf{Why probing avoids decoding losses.}
\Cref{thm:dg-miss} formalizes frequency bias; search errors and
exposure bias add the same pressure by pruning long-tail candidates
and compounding early decoding choices
\citep{stahlberg2019nmt,wiseman2016sequence,ranzato2015sequence,bengio2015scheduled}.
SPECTRA avoids these list-formation losses by scoring all categories
independently under the same context.

\textbf{Connection to discrete-choice preference modeling.}
SPECTRA's softmax readout (\cref{sec:method:primitive}) is the
multinomial-logit (MNL) parametrization of user preferences over
$C$: $\theta(c) \propto \exp(S(c)/\tau)$
\citep{luce1959individual,mcfadden1972conditional,train2009discrete}.
Under this lens, SPECTRA estimates MNL utilities $S$ from a
fine-tuned LLM via probing, and \cref{thm:ranking-advantage} is a
bounded-estimation-error guarantee on the implied preference
ranking. This grounds SPECTRA in classical discrete-choice theory
and connects to multi-interest user-representation work that targets
distributional rather than point-estimate preferences
\citep{shi2023everyone,cen2020controllable}.

\textbf{How the conditions are checked.}
For each user, the largest absolute category error gives
$\varepsilon$. A TV bound derived from JS provides an alternative
(\cref{sec:appendix:proofs:lemma2}). The ranking gap comes from
$\theta_*$, and comparing $L$ with $G^*$ identifies a DG miss. The main experiments report aggregate evidence---head/tail
bias, calibration depth, and fairness differential---consistent with
the theorem's predicted regime.

\textbf{Theoretical extensions.}
\Cref{thm:ranking-advantage} extends to population-expectation and
finite-sample confidence-interval forms
(\cref{sec:appendix:ranking-advantage:population}). An alternative
formulation in terms of total-variation distance (combining
\cref{lem:dg-truncation} and \cref{lem:spectra-calibration}) appears
in \cref{sec:appendix:proofs:theorem}.

\section{Experiments}\label{sec:experiments}

We evaluate SPECTRA on distributional alignment, long-tail recovery
with cross-user fairness, downstream value, and robustness.
\textit{Alignment} (\cref{sec:exp_alignment}) measures fidelity to
the user's true category distribution. \textit{Long-tail recovery}
(\cref{sec:exp_longtail}) tests whether the gain comes from
head-to-tail mass redistribution and benefits tail-preference users.
\textit{Downstream value} (\cref{sec:exp_downstream}) reports
category-ranking quality on MovieLens, Yelp, and a deployed
short-video platform, while \textit{robustness}
(\cref{sec:exp_robustness}) checks user-activity levels, context
windows, and base language models.
The metrics mirror the support-mismatch mechanism in
\cref{sec:method:theory}: JS measures full-distribution fidelity,
entropy and bias measure tail coverage, and NDCG measures utility as a
ranking signal or prior.

\subsection{Setup}\label{sec:exp_setup}

We describe datasets, models, and metrics used below; full
configurations and additional baselines are in
\cref{sec:appendix:e1_e5}.

\textbf{Datasets.} \textit{MovieLens}~\citep{harper2015movielens}
($K{=}19$ genres, $n{=}19{,}293$ users); \textit{Yelp} ($K_1{=}26$ L1
domains, $K{=}1{,}311$ L1$\times$L2 categories, $n{=}9{,}755$);
\textit{Short-Video}, a short-video platform dataset ($K{=}78$ categories, $50$M
interactions, $n{=}16$k).

\textbf{Models.} Qwen3-8B~\citep{qwen3technicalreport} fine-tuned
with LoRA~\citep{hu2022lora} under our CPT$+$SFT procedure; the
appendix repeats the same setup with
DeepSeek-R1-Distill-Llama-8B~\citep{deepseekai2025deepseekr1incentivizingreasoningcapability}.

\textbf{SPECTRA variants.} \textit{Likelihood-based probing}
(main-text default), \textit{generative classification}, and a
\textit{hierarchical variant} for large category sets, used on Yelp
L1$\times$L2 and Short-Video.

\textbf{Baselines.} \textit{Direct generation} from the same
fine-tuned LLM, \textit{Qwen3-Reranker}~\citep{qwen3embedding}, the
deployed production short-video ranker, and SPECTRA combined with
classical recommendation backbones
\textit{SASRec}~\citep{kang2018self} and
\textit{GRU4Rec}~\citep{hidasi2015session}; fusion results in
\cref{sec:appendix:downstream_utility}.

\textbf{Metrics.} We report NDCG@$k$ ($k\in\{5,10,20\}$) and
Jensen-Shannon divergence to the empirical category distribution.
Long-tail behavior is summarized by top-$k$ exposure
entropy ($k\in\{1,3\}$), per-bucket mass, stratified expected
calibration error (ECE), and mean absolute per-category bias under
a head/middle/tail partition (categories sorted by global frequency
into thirds).
Significance uses paired Wilcoxon with Holm-Bonferroni at
$\alpha{=}0.05$, and uncertainty uses bootstrap $95\%$ confidence
intervals from $1{,}000$ user-level resamples.

\textbf{Relation to \cref{thm:ranking-advantage}.} The results below
report overall distributional and per-user JS metrics, not the
per-user conditions of the theorem directly; the findings are
consistent with the theorem's prediction that probing reaches
categories DG skips.

\subsection{Distributional alignment}\label{sec:exp_alignment}

\begin{figure}[t]
  \centering
  \includegraphics[width=\linewidth]{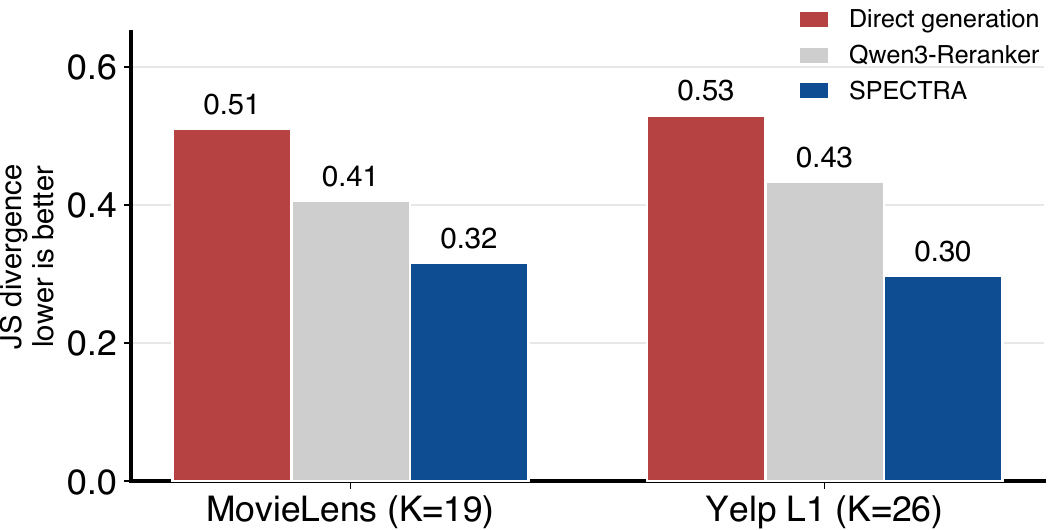}
  \vspace{-5pt}
  \caption{Jensen-Shannon divergence to the empirical category
  distribution on MovieLens ($K{=}19$) and Yelp L1 ($K{=}26$).
  Bars compare DG, Qwen3-Reranker, and SPECTRA. Numerical values
  with bootstrap $95\%$ confidence intervals (CIs) are in
  \cref{tab:alignment_js_appendix}.}
  \label{fig:alignment_summary}
\vspace{-10pt}
\end{figure}

We test whether SPECTRA's category distribution sits closer to the
user's true preferences than DG's.
\Cref{fig:alignment_summary} shows Jensen-Shannon divergence to the
empirical preference distribution on both MovieLens and Yelp L1
under the anchor configuration. On MovieLens, SPECTRA reduces JS
divergence by $0.19$ over DG ($38\%$ relative)
and by $0.09$ over the Qwen3-Reranker. On Yelp L1, the
corresponding reductions are larger at $0.23$ ($44\%$) over
DG and $0.14$ over Qwen3-Reranker. The larger Yelp gain tracks
its sharper head-tail asymmetry (\cref{sec:exp_longtail}). Numerical values with
bootstrap $95\%$ CIs are in
\cref{tab:alignment_js_appendix}~(\cref{sec:appendix:e1_e5:alignment_js}).

\subsection{Long-tail recovery with cross-user fairness}\label{sec:exp_longtail}

We examine how SPECTRA allocates probability across head and tail
categories, then compare the JS gains for head-leaning and
tail-leaning users.

\textbf{Head-tail mass redistribution.}
\Cref{tab:longtail_redistribution} reports per-bucket mass and head-bucket
calibration on both datasets. SPECTRA allocates 1.7 percentage
points more mass to the tail than DG on MovieLens and 0.1 points
more on Yelp. On MovieLens, both methods allocate more than the
observed tail share of $9.7\%$. Head-stratified ECE is $9.4\times$
the aggregate ECE on MovieLens and $2.3\times$ on Yelp.

\begin{table}[H]
\centering
\caption{\textit{Head-tail mass redistribution.} Anchor
configuration. \textit{(A)} probability mass on the tail bucket.
\textit{(B)} head-bucket ECE and its ratio to aggregate ECE.
\textit{(C)} mean absolute per-category bias on the head bucket
($\downarrow$).}
\label{tab:longtail_redistribution}
\vspace{-5pt}
\small
\setlength{\tabcolsep}{3pt}
\resizebox{\columnwidth}{!}{%
\begin{tabular}{@{}l c rr rr rr@{}}
\toprule
& & \multicolumn{2}{c}{A: tail mass}
  & \multicolumn{2}{c}{B: head ECE}
  & \multicolumn{2}{c}{C: head bias} \\
\cmidrule(lr){3-4}\cmidrule(lr){5-6}\cmidrule(lr){7-8}
\textit{Dataset} & $K$
  & SPECTRA & DG
  & value & ratio
  & SPECTRA & DG \\
\midrule
MovieLens & 19 & $\mathbf{0.239}$ & $0.222$ & $0.041$ & $9.4\times$ & $\mathbf{0.040}$ & $0.068$ \\
Yelp L1   & 26 & $\mathbf{0.278}$ & $0.277$ & $0.066$ & $2.3\times$ & $\mathbf{0.087}$ & $0.142$ \\
\bottomrule
\end{tabular}}
\end{table}

\textbf{Top-$k$ exposure entropy.}
SPECTRA also exposes a more diverse set of categories at each
user's top picks. On MovieLens, top-$1$ exposure entropy is $1.55$
for SPECTRA
versus $0.92$ for DG ($+68\%$), and top-$3$ is $3.01$ versus $2.44$
($+23\%$).

\begin{figure}[t]
  \centering
  \includegraphics[width=\linewidth]{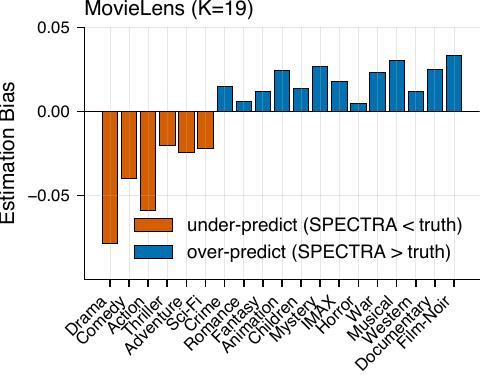}
  \vspace{-5pt}
  \caption{Per-category bias $\theta_S - \theta_*$ on MovieLens
  ($K{=}19$), sorted by true frequency. Orange marks head
  under-prediction ($\theta_S < \theta_*$); blue marks tail
  over-prediction ($\theta_S > \theta_*$). Yelp L1 panel in
  \cref{fig:per_cluster_bias_yelp_appendix}.}
  \label{fig:per_cluster_bias}
  \vspace{-10pt}
\end{figure}

\textbf{Per-category bias structure.}
\Cref{fig:per_cluster_bias} visualizes the asymmetry at the category
level. SPECTRA systematically under-predicts head categories (orange
bars, $\theta_S - \theta_* < 0$) and over-predicts tail categories
(blue bars, $\theta_S - \theta_* > 0$). On MovieLens the largest
under-predictions are on \textit{Drama} ($-7.9$\%),
\textit{Action} ($-5.9$\%), and \textit{Comedy} ($-4.0$\%); the
reallocated mass is spread across $13$ smaller-frequency categories.
On Yelp the asymmetry is sharper at the category level
(\cref{fig:per_cluster_bias_yelp_appendix}). Under paired Wilcoxon
with Holm-Bonferroni correction at $\alpha{=}0.05$, SPECTRA wins on
$17$ of $19$ MovieLens categories and $11$ of $26$ Yelp L1
categories; the two MovieLens losses (\textit{Musical},
\textit{Film-Noir}) both have truth mass below $1.5\%$.

\begin{figure}[t]
  \centering
  \includegraphics[width=\linewidth]{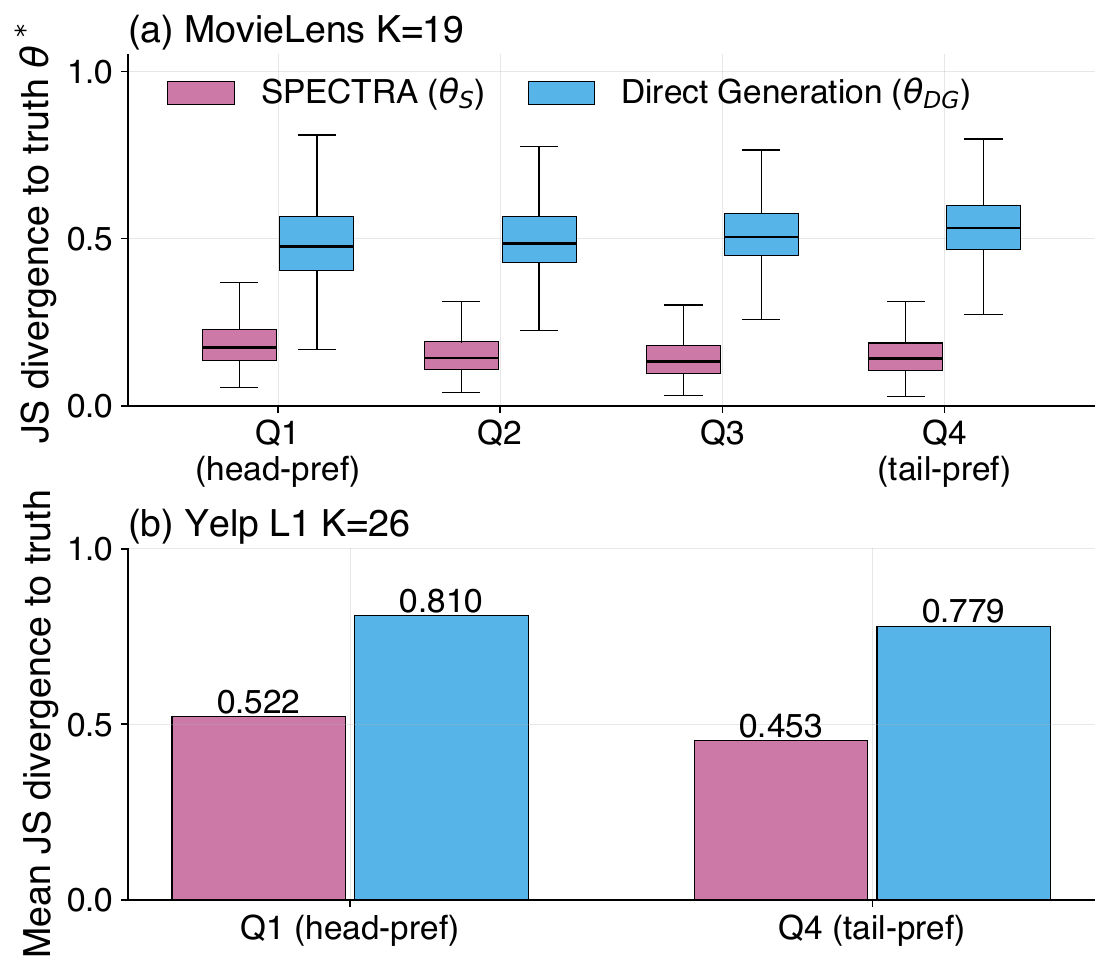}
  \vspace{-5pt}
  \caption{Per-user JS divergence to truth by tail-preference
  quartile. Quartiles are sorted by share of true future mass on
  tail categories. Q1 = head-leaning users, Q4 = tail-leaning users.
  (a) MovieLens, $K{=}19$. (b) Yelp L1, $K{=}26$, with Q1 and Q4 shown.
  Estimates and bootstrap $95\%$ CIs for Q1 and Q4 are in
  \cref{tab:user_level_fairness_appendix}.}
  \label{fig:user-level-fairness}
  \vspace{-10pt}
\end{figure}

\textbf{Cross-user fairness.}
Distributional gains can in principle be concentrated on
head-preference users while leaving tail-preference users behind.
\Cref{fig:user-level-fairness} compares users by the share of their
future interactions in the least frequent third of categories.
SPECTRA reduces mean per-user JS divergence in both the head-leaning
(Q1) and tail-leaning (Q4) groups on both datasets. The reduction in
Q4 is $1.28\times$ that in Q1 on MovieLens and $1.13\times$ on Yelp L1
(paired Wilcoxon $p<10^{-55}$ for Yelp Q4, $n{=}330$).
The Q1 and Q4 estimates and bootstrap $95\%$ CIs are in
\cref{tab:user_level_fairness_appendix}~(\cref{sec:appendix:e1_e5:fairness}).

\subsection{Downstream application value}\label{sec:exp_downstream}

\begin{figure}[t]
  \centering
  \includegraphics[width=\linewidth]{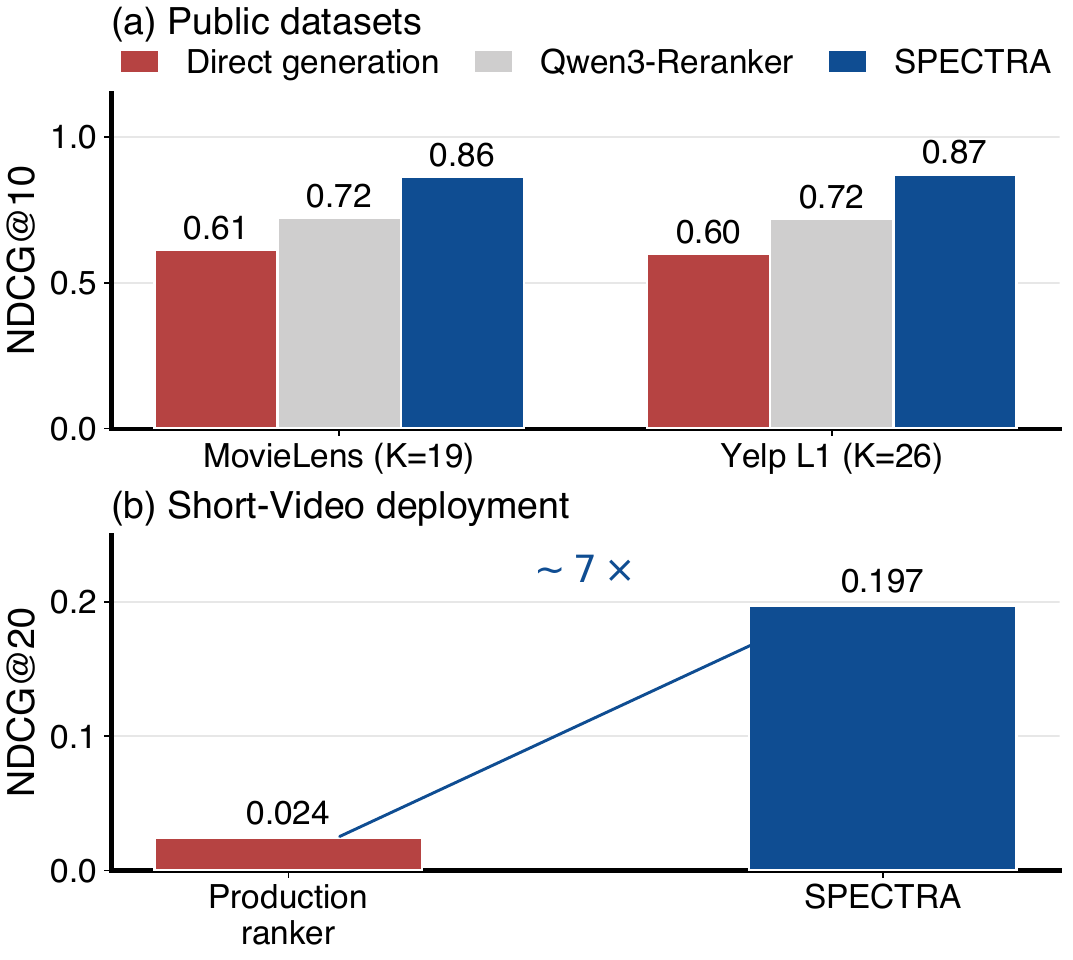}
  \vspace{-5pt}
  \caption{(a) Category-ranking NDCG@$10$ on MovieLens and Yelp L1
  ($\uparrow$). (b) Long-tail NDCG@$20$ gap on the deployed
  short-video platform. The annotated gain is
  $(0.197-0.024)/0.024$. Public-dataset estimates and bootstrap $95\%$
  CIs are in \cref{tab:downstream_ndcg_appendix}.}
  \label{fig:downstream_summary}
\vspace{-10pt}
\end{figure}

We next test whether ranking and discovery built on top of SPECTRA
also improve. SPECTRA delivers a $41\%$ relative NDCG@$10$ lift
over DG on MovieLens, and on Yelp's large category vocabulary it
lifts NDCG@$1$ from $0.008$ to $0.980$, since DG's sharply peaked
top-$1$ list rarely contains the true category.
\Cref{fig:downstream_summary} shows the cross-dataset NDCG pattern;
per-$k$ numerical detail is in
\cref{tab:downstream_ndcg_appendix}~(\cref{sec:appendix:e1_e5:downstream_ndcg}).

\textbf{Industrial-scale deployment.}
On the deployed short-video platform, SPECTRA's NDCG@$10$ on the
tail-category subset reaches $0.203$ against the production
Transformer-based behavior-sequence ranker's $0.034$. On the
long-tail slice with a $14$-day history and $14$-day prediction
horizon ($K{=}78$ categories), SPECTRA reaches
NDCG@$20{=}0.197$ against the production baseline's $0.024$,
a $721\%$ relative increase on this tail slice using the displayed
values. The comparison uses logged data and is restricted to
category-level tail ranking. The production system is optimized
for head items. \Cref{sec:appendix:robustness:industrial} reports
a separate cohort-level JS comparison.

\textbf{SPECTRA as a category prior for classical recommenders.}
We also test whether SPECTRA can boost classical sequential
recommenders by fusing its category distribution as a prior on top
of SASRec~\citep{kang2018self} and
GRU4Rec~\citep{hidasi2015session} on MovieLens. SPECTRA gives a
significant lift on the mid-popularity item slice ($+40\%$ relative on
SASRec NDCG@$10$, $+62\%$ on GRU4Rec). This is an item-level
per-slice result, distinct from the category-NDCG comparison above.
The overall NDCG@$10$ lift is small but positive. Full numbers in
\cref{sec:appendix:downstream_utility}.

\subsection{Robustness}\label{sec:exp_robustness}

We re-evaluate the fairness gain across $22$ cells defined by
user-activity strata and context-window settings on both datasets.
Every cell shows SPECTRA reducing per-user JS by at least $0.23$
relative to DG (paired-Wilcoxon $p<10^{-20}$); per-stratum ranges
are in \cref{tab:strata_summary_appendix}.

The appendix reports further diagnostics.
\textit{Taxonomy granularity.} \Cref{sec:appendix:robustness:taxonomy}
reports probe scores after coarsening the MovieLens taxonomy from
$K{=}19$ to $15$ and $10$ categories.
\textit{Prompt and verbalizer choice.} The variation is within $\pm
0.02$ NDCG (\cref{sec:appendix:robustness:prompt}).
\textit{Top-$1$ ECE.} The error stays below $0.005$ across cells
(\cref{sec:appendix:robustness:calibration_diag}).
\textit{Wall-clock latency.} Per-query latency scales linearly with
$K$ but is parallelizable across categories
(\cref{sec:appendix:robustness:latency}).
\textit{Base-model robustness.} DeepSeek-R1-Distill-Llama-8B
preserves the probe's NDCG and JS advantage over DG on MovieLens
(\cref{sec:appendix:e1_e5:deepseek}).

\begin{figure}[t]
  \centering
  \includegraphics[width=\linewidth]{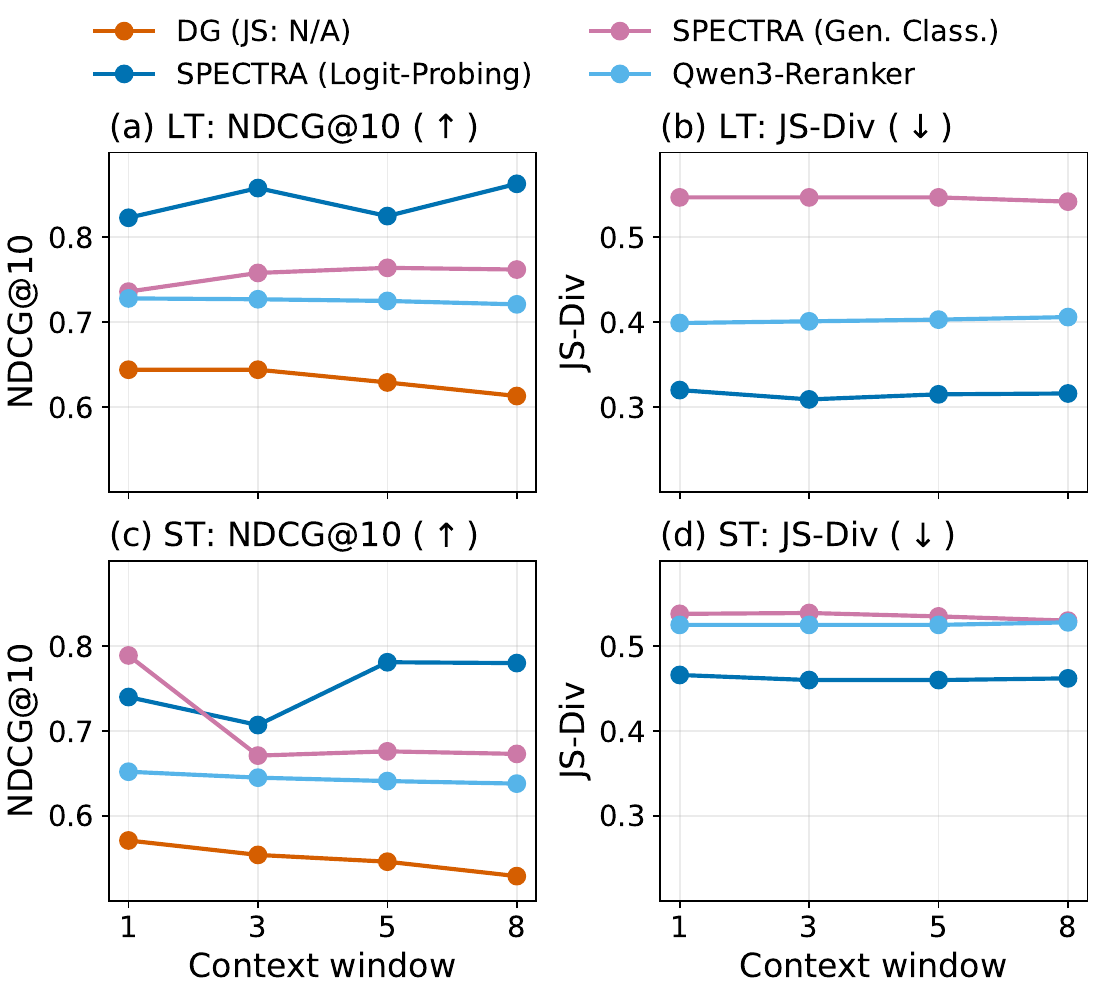}
  \vspace{-5pt}
  \caption{Context-window robustness on MovieLens (Qwen3-8B).
  NDCG@$10$ ($\uparrow$, left) and JS divergence ($\downarrow$,
  right) for SPECTRA's two scoring variants, DG, and
  Qwen3-Reranker-8B across context windows $\{1,3,5,8\}$. Top row is
  the long-term horizon; bottom row is short-term.}
  \label{fig:movielens_trends}
\vspace{-10pt}
\end{figure}

\textit{Context-window scaling.}
\Cref{fig:movielens_trends} plots NDCG@$10$ and JS divergence across
context windows $\{1,3,5,8\}$ for SPECTRA's two scoring variants
(logit-probing and generative classification), DG, and
the Qwen3-Reranker baseline on MovieLens. The likelihood probe
outperforms the reranker on both metrics throughout. The generative variant has
higher long-term JS than the reranker. Figure~\ref{fig:movielens_trends}
reports NDCG for DG. On
Yelp the per-user JS reduction widens $45\%$ as context grows, from
$0.250$ at context window $1$ to $0.362$ at context window $8$, a
pattern absent on MovieLens. The mechanism is dataset-shape dependent. MovieLens has a
multi-genre head that amplifies asymmetry with longer context, while
Yelp's single dominant head (Food and Restaurants at $64\%$) anchors
more tightly as context grows, leaving more room for SPECTRA to
recover the tail.

\subsection{\texorpdfstring{\revision{Classification Models and Readouts}}{Classification Models and Readouts}}
\label{sec:exp_readout_comparison}

\revision{We compared SPECTRA with existing LLM classifiers and readouts
on a short-video task (\cref{tab:classification_readouts}). The
outcome is binary for each user and category: whether the user
views that category during the prediction window. Several
categories can be positive for the same user.}
\begin{itemize}
\item \revision{\textbf{LLM-based classifiers.}
Jev~\citep{almeida2026jev} is a commercial decision model that returns
probabilities over predefined answers. Kev-4B~\citep{palmer2026kev4b}
is an open model that returns the same type of output. We use both
without fine-tuning on our interaction data.}
\item \revision{\textbf{LLM readout methods.}
We use the same fine-tuned LLM for each method. Complete-answer
likelihood~\citep{shysheya2025jolt} scores all tokens in each candidate
answer and normalizes across answers. Numerical
likelihood~\citep{requeima2024llmp} scores numerical answers.
Answer-code rotation~\citep{zhang2026anyjev} reassigns codes to the
answer options and averages their scores across assignments.}
\end{itemize}

\revision{On the shared test set, SPECTRA had the lowest log loss
(0.312) and highest AUC (0.936). Its log loss was 51\% lower than
Kev-4B's and 33--39\% lower than the other readouts'.
Jev's score comes from a separate validation panel.}

\begin{table}[t]
  \centering
  \small
  \revision{%
  \begin{tabular*}{\linewidth}{@{\extracolsep{\fill}}lcc@{}}
    \toprule
    & Log loss $\downarrow$ & AUC $\uparrow$ \\
    \midrule
    SPECTRA~{\footnotesize(\cref{sec:method:primitive})} & \textbf{0.312} & \textbf{0.936} \\
    \midrule
    \multicolumn{3}{@{}l@{}}{\textit{Other LLM-based classification models}} \\
    Jev\textsuperscript{1}~{\footnotesize \citep{almeida2026jev}} & 0.422 & 0.874 \\
    Kev-4B~{\footnotesize \citep{palmer2026kev4b}} & 0.640 & 0.608 \\
    \midrule
    \multicolumn{3}{@{}l@{}}{\textit{Other readouts of the same fine-tuned LLM}} \\
    \begin{tabular}[t]{@{}l@{}}Answer-code rotation\\[-1pt]{\footnotesize \citep{zhang2026anyjev}}\end{tabular} & 0.465 & 0.840 \\
    \begin{tabular}[t]{@{}l@{}}Complete-answer likelihood\\[-1pt]{\footnotesize \citep{shysheya2025jolt}}\end{tabular} & 0.508 & 0.784 \\
    \begin{tabular}[t]{@{}l@{}}Numerical likelihood\\[-1pt]{\footnotesize \citep{requeima2024llmp}}\end{tabular} & 0.515 & 0.790 \\
    \bottomrule
  \end{tabular*}}
  \caption{\revision{Predicting whether a user views a category on the
  short-video platform (78 categories). SPECTRA, Kev-4B, and the
  three other readouts use the same 500-user test panel after
  calibration on held-out validation users.
  \textsuperscript{1}Jev is evaluated separately by cross-fitting on
  261 validation users.}}
  \label{tab:classification_readouts}
\end{table}

\section{Conclusion}
\label{sec:conclusion}

We presented SPECTRA, a logit-probing method that recovers
category-level user preference distributions from a fine-tuned LLM's
softmax. Across DG and recommender baselines, SPECTRA
improves distributional fidelity and long-tail recovery, while a
ranking-advantage theorem characterizes when this gain
is provable. Named-category distributions make the learned user state
inspectable and reusable, offering a compact path toward structured
preference modeling with LMs.

\section*{Limitations}
\label{sec:limitations}

This study focuses on category-level distribution readout under fixed
taxonomies and offline evaluation. Natural extensions include broader
coverage of decoding strategies, prompt templates, random seeds, and
public or industrial domains.
\revision{Future work could vary the mix of interaction data used for
fine-tuning~\citep{zhang2026pickmixturesamplescombinatorial} to test how it affects recovery of
long-tail preferences.}
Category-history priors remain useful
stationarity diagnostics when the goal is forecasting tomorrow's
category mix; SPECTRA instead asks whether an aligned LLM exposes a
calibrated internal preference distribution. Downstream work should
evaluate how cached category distributions affect full recommender
objectives and user-facing outcomes over time.

\section*{Acknowledgments}
We thank the anonymous reviewers for their valuable feedback. We also thank Siyuan Peng, Yan Li, and Zhenheng Yang for helpful internal discussions.

\bibliography{custom}

\appendix
\crefalias{section}{appendix}
\crefalias{subsection}{appendix}
\crefalias{subsubsection}{appendix}

\section{Reproducibility checklist}
\label{sec:appendix:repro}

\subsection{ARR reproducibility checklist}
\label{sec:appendix:repro:checklist}

We follow the ARR June 2026 reproducibility checklist. Items are answered
below; the corresponding manuscript locations are cross-referenced where
applicable.

\textbf{(C1) Description of computing infrastructure.} The adaptation and
GPU inference jobs summarized in \cref{sec:appendix:repro:compute}
used 8$\times$NVIDIA
L40S (48\,GB) GPUs. Each L40S supplies $\sim$91 TFLOPS BF16 throughput; jobs
used PyTorch 2.x, transformers, PEFT, and (for inference) vLLM. CPU
re-analysis (calibration depth, fairness stratification, prior ablation, bootstrap
confidence intervals) used a single 16-core CPU partition node. We list per-job
wall-clock estimates in \cref{sec:appendix:repro:compute}.

\textbf{(C2) Number of parameters in each model.} The base model is
Qwen3-8B~\citep{qwen3technicalreport} ($\sim$8.2B parameters). The LoRA
adapter for the CPT+SFT stage on Yelp Qwen3 occupies $\sim$61\,MB of disk
($\sim$31M trainable parameters at our LoRA configuration); MovieLens
Qwen3 LoRA $\sim$174\,MB ($\sim$88M trainable parameters). The base-model
robustness panel uses
DeepSeek-R1-Distill-Llama-8B~\citep{deepseekai2025deepseekr1incentivizingreasoningcapability}
($\sim$8B parameters) with separately trained LoRA adapters of comparable
size.

\textbf{(C3) Average runtime for each model or for each algorithm.}
Per-dataset inference runtimes on 8$\times$L40S (data-parallel):
\begin{itemize}
  \item MovieLens, $K{=}19$, main evaluation cohort: $\sim$10--14\,h
        wall-clock for likelihood probing with prefill caching, including
        all four context-window settings.
  \item Yelp L1, $K{=}26$, main evaluation cohort: $\sim$10\,h
        wall-clock. The
        hierarchical L2 pass over $\sim$26 L1 branches reuses the L1 prefill
        cache.
  \item Short-video, $K{=}78$: aggregate metrics only; reported in
        \cref{sec:appendix:robustness:industrial}.
\end{itemize}
LoRA SFT training (Qwen3-8B Yelp) ran
for $15{,}000$ steps with effective batch size $6\times 2\times 8 = 96$ on
$8$ GPUs; total wall-clock $\sim$1 day with a cosine schedule and $10\%$
warmup.

\textbf{(C4) Probe settings.}
The probe has three exposed hyperparameters: the softmax temperature
$\tau$, the affirmative verbalizer set $V_+$, and the negative verbalizer
set $V_-$ (\cref{alg:spectra-likelihood}, mirrored in
\cref{sec:appendix:e1_e5:tables}). We did not sweep these; they were fixed
at the values inherited from the original LoRA SFT recipe
($\tau=1.0$; $V_+ = \{\texttt{Yes}, \texttt{Y}, \texttt{y}\}$;
$V_- = \{\texttt{No}, \texttt{N}, \texttt{n}\}$).
The hierarchical
probe has one additional setting (the L1 branching depth $B$, fixed at
$B=K_{1}$, i.e., all 26 L1 branches were expanded for Yelp).
\Cref{tab:compute_summary} reports wall-clock times.

\textbf{(C5) Bounds for each reported result.} Bootstrap $95\%$ confidence
intervals, where shown, are computed from $1{,}000$ user-level resamples
of the eval cohort (paired-resample design when comparing methods; uniform
resample when reporting a single method's mean). Paired-Wilcoxon signed-rank
$p$-values are reported for cross-method comparisons.

\textbf{(C6) Description of train/validation/test split.} On MovieLens and Yelp,
the first $80\%$ of each user's chronological interaction history
is used for CPT+SFT and the remaining $20\%$ for evaluation. Users may
appear in both portions. Held-out future interactions are excluded from
the context and adaptation targets for the corresponding example.
The short-video ranking and incidence evaluations are described in
\cref{sec:exp_downstream,sec:exp_readout_comparison}, and the cohort-level
short-video JS comparison in \cref{sec:appendix:robustness:industrial}.

\textbf{(C7) Description of statistical significance testing.} Comparisons
between SPECTRA and DG use paired Wilcoxon signed-rank tests at the
per-user level, with $n=19{,}293$ on MovieLens and $n=9{,}755$ on Yelp.
We additionally report bootstrap $95\%$ CIs (\cref{sec:appendix:e1_e5}) on
all headline metrics. The downstream item-utility demo
(\cref{sec:appendix:downstream_utility}) uses paired Wilcoxon at
$n_{\mathrm{paired}}=17{,}498$ (the user subset with non-empty
test-window).

\textbf{(C8) Hyperparameters used during training (LoRA SFT).}
\begin{itemize}
  \item Base model: Qwen3-8B (alternate base for E5: DeepSeek-R1-Distill-Llama-8B).
  \item Optimization: AdamW with cosine schedule and $10\%$ warmup.
  \item Learning rate: $8\!\times\!10^{-5}$.
  \item Per-GPU batch size: $6$ samples; gradient-accumulation steps: $2$;
        GPUs: $8$. Effective batch size $= 6 \times 2 \times 8 = 96$ samples.
  \item Training steps: $15{,}000$ (Yelp); MovieLens trained to comparable
        wall-clock and converged earlier given the smaller corpus.
  \item Random seed: $42$ (single seed; we note this as a limitation in
        \cref{sec:limitations}).
  \item LoRA framework: \texttt{llama-factory} on top of PEFT
        \citep{hu2022lora}; final validation loss $0.1933$ on Yelp.
  \item LoRA hyperparameters (PEFT \texttt{adapter\_config.json},
        \texttt{peft\_type}=\texttt{LORA}, \texttt{task\_type}=\texttt{CAUSAL\_LM}):
        $r{=}32$; \texttt{lora\_alpha}$=64$ (scaling $\alpha/r = 2$);
        \texttt{lora\_dropout}$=0.1$; \texttt{bias}=\texttt{none};
        \texttt{target\_modules} = \texttt{q\_proj}, \texttt{k\_proj},
        \texttt{v\_proj}, \texttt{o\_proj}, \texttt{gate\_proj},
        \texttt{up\_proj}, \texttt{down\_proj} (all attention projections
        plus all MLP projections); \texttt{use\_rslora}=\texttt{False};
        \texttt{use\_dora}=\texttt{False}; \texttt{init\_lora\_weights}=\texttt{True}.
  \item Continued pretraining (CPT) used the domain corpus (textualized user
        metadata, item metadata, and serialized interaction traces) under
        the same optimizer/schedule before SFT.
\end{itemize}

\textbf{(C9) Data used.}
\begin{itemize}
  \item \emph{MovieLens.} The MovieLens public release
        \citep{harper2015movielens}, accessed under the MovieLens
        terms of use. We use the $K=19$ flat genre taxonomy. The CPT/SFT
        corpus is built from the public ratings + movie-metadata files. The
        headline public-dataset evaluation uses $19{,}293$ MovieLens users;
        stage-wise and robustness analyses may use approximately $20{,}000$
        probe records because they materialize one or more context windows per
        user.
  \item \emph{Yelp.} The Yelp Open Dataset (academic license; restricted to
        non-commercial research use), with a curated L1$\leftrightarrow$L2
        hierarchy (\cref{sec:appendix:e1_e5:yelp_hierarchy}). $K_{1}=26$ L1 categories;
        $1{,}311$ L2 categories under the union of L1 branches.
  \item \emph{Short-video subset.} An observational interaction-log subset
        provided by an industry collaborator under a non-disclosure
        data-sharing agreement. The subset used for the cohort-level JS comparison comprises high-activity users
        ($\geq 21$ active days in the prior month) with $30$-day history and
        a $14$-day prediction window, $K=78$ categories. Only aggregate metrics
        are reported (\cref{sec:appendix:robustness:industrial}). No raw
        per-user records are released.
\end{itemize}

\textbf{(C10) Codebase and license.} The companion repository linked in
\cref{sec:intro} contains the SPECTRA probing implementation, prompt
templates, evaluation metrics, and runnable examples under the MIT
license. The proprietary short-video corpus is covered by a
data-sharing agreement.

\subsection{Compute and wall-clock summary}
\label{sec:appendix:repro:compute}

\Cref{tab:compute_summary} summarizes wall-clock times for inference and re-analysis. Latency tables for the per-query cost of probing versus DG
appear in \cref{sec:appendix:robustness:latency}.

\begin{table}[h]
\centering
\caption{Per-job wall-clock for the inference and re-analysis runs reported
in this submission. Training wall-clock for the LoRA adapters used as
inputs is reported in \cref{sec:appendix:repro:checklist} (C3).}
\label{tab:compute_summary}
\small
\resizebox{\columnwidth}{!}{%
\begin{tabular}{lcc}
\toprule
\textbf{Job} & \textbf{Hardware} & \textbf{Wall-clock} \\
\midrule
MovieLens Qwen3-8B (E1--E4) & 8$\times$L40S & $\sim$10--14\,h \\
Yelp Qwen3-8B (L1$+$L2)     & 8$\times$L40S & $\sim$10\,h \\
DeepSeek base-model (E5)    & 8$\times$L40S & $\sim$12\,h \\
Downstream item-utility demo & 16-core CPU & $\sim$4\,min \\
Calibration / fairness re-analysis & 16-core CPU & $\sim$30\,min \\
\bottomrule
\end{tabular}}
\end{table}

\subsection{Direct-generation decoder}
\label{sec:appendix:repro:dg}
The DG decoder used throughout the main experiments is a
\emph{uniform-on-top-$5$} renormalization of the LM's argmax-selected
$k=5$-category list, with $k$ matched to a typical ranked-list cut-off.
This is the standard cluster-level analog of autoregressive ranking
output, since DG produces a ranked list rather than a distribution
and the uniform conversion weights the five selected categories equally.
The truncation bound in \cref{lem:dg-truncation} holds for any weighting
supported on the selected list.

\subsection{SPECTRA probing algorithms}
\label{sec:appendix:algorithm}

\begin{algorithm}[H]
\caption{SPECTRA likelihood-based probing.}
\label{alg:spectra-likelihood}
\begin{algorithmic}[1]
\Require LLM $\mathcal{M}$; context $X_{1:t}$; category set
$C = \{c_1, \ldots, c_K\}$; affirmative tokens $V_+$, negative
tokens $V_-$; temperature $\tau$.
\Ensure $\theta_S$ (probability distribution over $C$).
\State $S \leftarrow \mathbf{0} \in \mathbb{R}^K$.
\For{$j = 1$ to $K$}
  \State $\mathbf{z}_j \leftarrow \mathcal{M}\bigl(\pi_{\text{probe}}(X_{1:t}, c_j)\bigr)$
         \Comment{next-token logits}
  \State $S_j \leftarrow \tfrac{1}{|V_+|}\!\sum_{v \in V_+}\!\mathbf{z}_j[v]
                        - \tfrac{1}{|V_-|}\!\sum_{v \in V_-}\!\mathbf{z}_j[v]$
\EndFor
\State \textbf{return} $\theta_S \leftarrow \operatorname{softmax}(S / \tau)$.
\end{algorithmic}
\end{algorithm}

\begin{algorithm}[H]
\caption{SPECTRA generative classification.}
\label{alg:spectra-generative}
\begin{algorithmic}[1]
\Require LLM $\mathcal{M}$; context $X_{1:t}$; category set
$C = \{c_1, \ldots, c_K\}$; answer-token mapping
$\{c_j \mapsto v_j\}_{j=1}^{K}$ (e.g.\ $v_j \in \{\texttt{A},
\texttt{B}, \dots\}$); temperature $\tau$.
\Ensure $\theta_S$ (probability distribution over $C$).
\State $\mathbf{z} \leftarrow \mathcal{M}\bigl(\pi_{\text{gen}}(X_{1:t}, C)\bigr)$
       \Comment{single next-token logit vector}
\For{$j = 1$ to $K$}
  \State $S_j \leftarrow \mathbf{z}[v_j]$
\EndFor
\State \textbf{return} $\theta_S \leftarrow \operatorname{softmax}(S / \tau)$.
\end{algorithmic}
\end{algorithm}

\section{Detailed E1--E5 results and DeepSeek base-model robustness}
\label{sec:appendix:e1_e5}

This appendix expands the main-text headline tables (\cref{sec:experiments})
with numerical summaries, context-window results, and the
DeepSeek-R1-Distill-Llama-8B base-model comparison (E5).

\subsection{Distributional alignment numerical values}
\label{sec:appendix:e1_e5:alignment_js}

\Cref{tab:alignment_js_appendix} reports the numerical values
visualized in \cref{fig:alignment_summary}, with bootstrap $95\%$ CI
half-widths from $1{,}000$ user-level resamples.

\begin{table}[t]
\centering
\caption{\textit{Distributional alignment (numerical detail for
\cref{fig:alignment_summary}).} Jensen-Shannon divergence to the
empirical category distribution, in bits ($\downarrow$). Anchor
configuration. Subscripts are bootstrap $95\%$ CI half-widths from
$1{,}000$ user-level resamples ($n{=}19{,}293$ MovieLens,
$n{=}9{,}755$ Yelp L1).}
\label{tab:alignment_js_appendix}
\small
\resizebox{\columnwidth}{!}{%
\begin{tabular}{lcc}
\toprule
\textit{Method} & MovieLens (K{=}19) & Yelp L1 (K{=}26) \\
\midrule
Direct generation        & $0.510_{\pm.006}$ & $0.530_{\pm.008}$ \\
Qwen3-Reranker-8B        & $0.406_{\pm.006}$ & $0.434_{\pm.008}$ \\
SPECTRA (probe)          & $\mathbf{0.316}_{\pm.005}$ & $\mathbf{0.298}_{\pm.007}$ \\
\bottomrule
\end{tabular}}
\end{table}

\subsection{Downstream NDCG numerical values}
\label{sec:appendix:e1_e5:downstream_ndcg}

\Cref{tab:downstream_ndcg_appendix} reports the numerical values
visualized in \cref{fig:downstream_summary} (left panel), with
bootstrap $95\%$ CI half-widths from $1{,}000$ user-level resamples.

\begin{table}[t]
\centering
\caption{\textit{Category-ranking quality (numerical detail for
\cref{fig:downstream_summary}).} NDCG@$k$ ($\uparrow$) under the
anchor configuration. Subscripts are bootstrap $95\%$ CI half-widths
from $1{,}000$ user-level resamples.}
\label{tab:downstream_ndcg_appendix}
\small
\resizebox{\columnwidth}{!}{%
\begin{tabular}{lcc cc}
\toprule
& \multicolumn{2}{c}{MovieLens (K{=}19)} & \multicolumn{2}{c}{Yelp L1 (K{=}26)} \\
\cmidrule(lr){2-3}\cmidrule(lr){4-5}
\textit{Method} & NDCG@5 & NDCG@10 & NDCG@5 & NDCG@10 \\
\midrule
Direct generation  & $0.530_{\pm.006}$ & $0.613_{\pm.006}$ & $0.512_{\pm.008}$ & $0.598_{\pm.008}$ \\
Qwen3-Reranker-8B  & $0.701_{\pm.006}$ & $0.721_{\pm.006}$ & $0.683_{\pm.008}$ & $0.720_{\pm.008}$ \\
SPECTRA (probe)    & $\mathbf{0.822}_{\pm.005}$ & $\mathbf{0.863}_{\pm.005}$ & $\mathbf{0.842}_{\pm.007}$ & $\mathbf{0.871}_{\pm.007}$ \\
\bottomrule
\end{tabular}}
\end{table}

\subsection{User-level fairness numerical values}
\label{sec:appendix:e1_e5:fairness}

\Cref{tab:user_level_fairness_appendix} reports the Q1-vs-Q4
fairness differential visualized in
\cref{fig:user-level-fairness}, with bootstrap $95\%$ CI half-widths
for Q1 and Q4 and a paired Wilcoxon $p$-value for Q4.

\begin{table}[t]
\centering
\caption{\textit{User-level fairness differential (numerical detail
for \cref{fig:user-level-fairness}).} $\Delta$ = mean per-user
$(\text{JS}_{\text{SPECTRA}} - \text{JS}_{\text{DG}})$ to truth, in
bits ($\downarrow$), paired per user. Subscripts are within-quartile
bootstrap $95\%$ CI half-widths. Yelp Q4 is a small cohort
($n{=}330$).}
\label{tab:user_level_fairness_appendix}
\small
\resizebox{\columnwidth}{!}{%
\begin{tabular}{l c rrr rr}
\toprule
\textit{Dataset} & $K$ & $\Delta_{\text{Q1 head}}$ & $\Delta_{\text{Q4 tail}}$ & \textit{ratio} Q4/Q1 & $n_{\text{Q4}}$ & $p_{\text{Q4}}$ \\
\midrule
MovieLens & 19 & $-0.300_{\pm.007}$ & $\mathbf{-0.383}_{\pm.009}$ & $\mathbf{1.28}$ & $4{,}823$ & $<10^{-100}$ \\
Yelp L1   & 26 & $-0.288_{\pm.005}$ & $\mathbf{-0.325}_{\pm.023}$ & $\mathbf{1.13}$ & $330$ & $7.62\!\times\!10^{-56}$ \\
\bottomrule
\end{tabular}}
\end{table}

\subsection{Stratified fairness gain (activity and context-window cells)}
\label{sec:appendix:e1_e5:strata}

\Cref{tab:strata_summary_appendix} reports the per-stratum range of
the per-user JS reduction (SPECTRA minus DG) across $22$ cells defined
by user-activity strata and context-window settings on both datasets.
Every cell holds at paired-Wilcoxon $p<10^{-20}$.

\begin{table}[t]
\centering
\caption{\textit{Per-stratum range of the per-user JS reduction
(SPECTRA minus DG), $\downarrow$.} Within-cell paired Wilcoxon
$p<10^{-20}$ throughout.}
\label{tab:strata_summary_appendix}
\small
\resizebox{\columnwidth}{!}{%
\begin{tabular}{l l rr}
\toprule
\textit{Dataset} & \textit{Stratum} & \textit{range} & \textit{cells wins} \\
\midrule
MovieLens & user activity (5 strata)    & $-0.325$ to $-0.373$ & 5/5 \\
MovieLens & context window (6 strata)   & $-0.342$ to $-0.376$ & 6/6 \\
Yelp L1   & user activity (5 strata)    & $-0.280$ to $-0.305$ & 5/5 \\
Yelp L1   & context window (6 strata)   & $-0.234$ to $-0.362$ & 6/6 \\
\bottomrule
\end{tabular}}
\end{table}

\subsection{Context-window results and experiment summary}
\label{sec:appendix:e1_e5:tables}

\Cref{fig:movielens_trends} compares the MovieLens methods across four
context windows of $\{1,3,5,8\}$ sessions for long- and short-term
prediction. The DeepSeek comparison appears in
\cref{sec:appendix:e1_e5:deepseek}. The Qwen3 likelihood probe
has lower JS than the reranker at each context window in the plot.

Headline numbers (Qwen3-8B, CW$=8$, LT) for orientation:
\begin{itemize}
  \item E1 (category-level NDCG, ranking the category vocabulary):
        SPECTRA NDCG@$10 = 0.863$ vs.\ DG $0.613$ vs.\
        Qwen3-Reranker-8B $0.721$.
  \item E3 (Yelp hierarchical, $K_{1}=26$ L1 with conditional L2):
        SPECTRA matches or exceeds DG on NDCG@$\{1,5,10\}$ in
        long-term and short-term predictions; see
        \cref{sec:appendix:e1_e5:yelp_hierarchy} for the L1$\times$L2
        structure detail.
  \item E4 (short-video long-tail recovery, $14$-day history,
        $14$-day prediction, $K=78$): SPECTRA NDCG@$20 = 0.197$ on the
        long-tail category subset vs.\ production baseline $0.024$
        (a $721\%$ relative increase on that slice, using the displayed values).
        \Cref{sec:appendix:robustness:industrial} reports a separate
        cohort-level JS comparison.
\end{itemize}

\subsection{Yelp L1 \texorpdfstring{$\times$}{x} L2 hierarchy detail}
\label{sec:appendix:e1_e5:yelp_hierarchy}

The Yelp category vocabulary is induced by a two-level taxonomy: $K_{1}=26$
L1 domains, with an L2 vocabulary conditional on each L1 parent. The L1
domain set, ordered by aggregate empirical mass, is:
``Food \& Restaurants'', ``Health \& Medical'', ``Shopping \& Retail'',
``Home \& Public Services'', ``Active Life, Sports \& Recreation'',
``Beauty \& Spas'', ``Arts, Entertainment \& Events'', ``Automotive'',
``Professional \& Financial Services'', ``Hotels \& Travel'',
``Education'', ``Nightlife \& Bars'', ``Real Estate'', ``Pets'',
``Religious \& Community'', ``Personal Services'',
``Event Planning \& Services'', ``Specialty Shops'',
``Local \& Public Services'', ``Internet \& Communications'',
``Cannabis Services'', ``Active Life \& Fitness'',
``Farms \& Ranches'', ``Community \& Government'',
``Home Maintenance'', and ``Specialty Vehicles''.

The L2 fan-out per L1 parent varies substantially: ``Food \& Restaurants''
has $231$ L2 children, ``Home \& Public Services'' has $167$, ``Shopping \& Retail''
has $144$, ``Arts, Entertainment \& Events'' has $143$,
``Health \& Medical'' has $143$, and
``Active Life, Sports \& Recreation'' has $109$ L2 children, while
several minority L1 domains have a single L2 child. The
union over the $26$ L1 parents yields the $K=1{,}311$ L2 category
vocabulary cited in the main text.

\textbf{How the hierarchical probe operates on this structure.} The
hierarchical probe (\cref{alg:spectra-likelihood}) factorizes
the joint
$\theta_S(c) = \theta_S(L_1(c))\cdot
\theta_S(L_2(c)\,\mid\,L_1(c))$. Step~1 issues
$K_{1}=26$ L1-level probing prompts (a single ``Yes/No'' likelihood probe
per L1 domain) and softmax-normalizes the resulting logit vector to
obtain an L1-level distribution $P_{L1}$. Step~2 conditions on the
L1 branches and issues a second round of L2-level probing prompts
within each branch. The reported Yelp evaluation expands all $26$
branches. The conditional probabilities
$P_{L2\mid L1}$ are softmax-normalized within the branch and combined
with $P_{L1}$ by the chain rule to give the joint over the full $K=1{,}311$
L2 category set. With KV-cache prefix reuse the L1 prompts share a single
user-history prefill; the L2 prompts share an L1-conditional prefill,
so the cost is $K_{1}\cdot C(\mathcal{M}, T_{\mathrm{suf}}) +
\sum_{j \in \mathrm{selected}} |L_2(j)| \cdot C(\mathcal{M},
T'_{\mathrm{suf}})$. With $B=K_{1}$ (all L1 branches expanded) and
average L2 fan-out $\bar{|L_2|} = 1311 / 26 \approx 50.4$, the
total prompt count is $26 + 1{,}311 = 1{,}337$. With full expansion,
prefix reuse reduces repeated prefill work. Expanding fewer branches
reduces the prompt count and leaves the conditional distribution within
unexpanded branches unspecified.

\subsection{E5 -- DeepSeek-R1-Distill-Llama-8B base-model robustness}
\label{sec:appendix:e1_e5:deepseek}

The base-model comparison checks whether SPECTRA's ranking and JS
advantage over DG carries over to a second backbone. We swap the base to
DeepSeek-R1-Distill-Llama-8B~\citep{deepseekai2025deepseekr1incentivizingreasoningcapability}
with a separately trained LoRA adapter (same recipe).
All other elements --- prompt template,
verbalizers, category vocabulary, eval cohort --- are held fixed.

Headline numbers on MovieLens (CW$=8$, LT):
\begin{itemize}
  \item SPECTRA logit probe: NDCG@$10 = 0.872$, JS $=0.367$ (DeepSeek
        base) vs.\ NDCG@$10 = 0.863$, JS $=0.316$ (Qwen3-8B base).
  \item DG: NDCG@$10 = 0.546$ (DeepSeek base) vs.\ $0.613$ (Qwen3-8B
        base).
\end{itemize}
The probe's advantage over DG (in both NDCG@$10$ and JS) holds under
the DeepSeek base across all four context windows and both prediction
horizons. The absolute JS values shift between bases (DeepSeek
$\sim 0.37$ vs.\ Qwen3 $\sim 0.32$), consistent with calibration depth
being base-specific, but the \emph{sign} and \emph{ranking} of
SPECTRA-vs-DG dominance are preserved.

\section{Calibration extras}
\label{sec:appendix:calibration}

This appendix expands the calibration-depth analysis in
\cref{sec:experiments} into the figures, scatter plots, and stratified
ECE tables that the page budget compresses in the main body.

\subsection{Cross-dataset reliability diagrams}
\label{sec:appendix:calibration:reliability}

\begin{figure}[t]
  \centering
  \includegraphics[width=\linewidth]{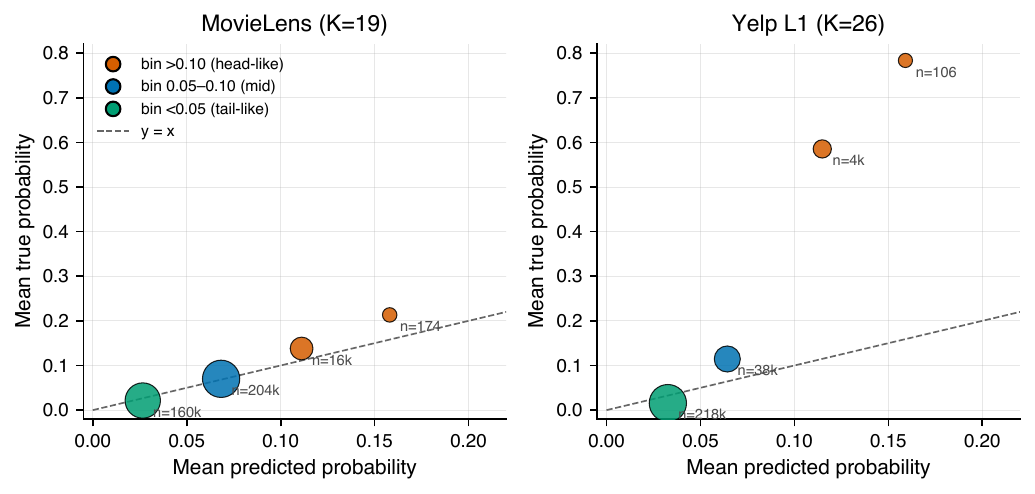}
  \caption{Soft-label reliability diagram on (left) MovieLens (K{=}19) and
  (right) Yelp L1 (K{=}26). Each point is a 0.05-wide bin of predicted
  category-level probability; the x-axis is the mean predicted probability in
  the bin and the y-axis is the mean true probability. Point size scales with
  bin count $n$. The highest predicted-probability bins systematically lie
  above the diagonal --- SPECTRA under-predicts the head categories.}
  \label{fig:reliability-diagram}
\end{figure}

\Cref{fig:reliability-diagram} bins predicted category probabilities
and compares each bin's mean predicted and observed shares. The highest
probability bins lie above the diagonal, so the observed shares are
higher than the predictions. This gap is larger on Yelp L1.

\subsection{Per-category confidence vs.\ frequency scatter}
\label{sec:appendix:calibration:scatter}

\begin{figure*}[t]
  \centering
  \includegraphics[width=\linewidth]{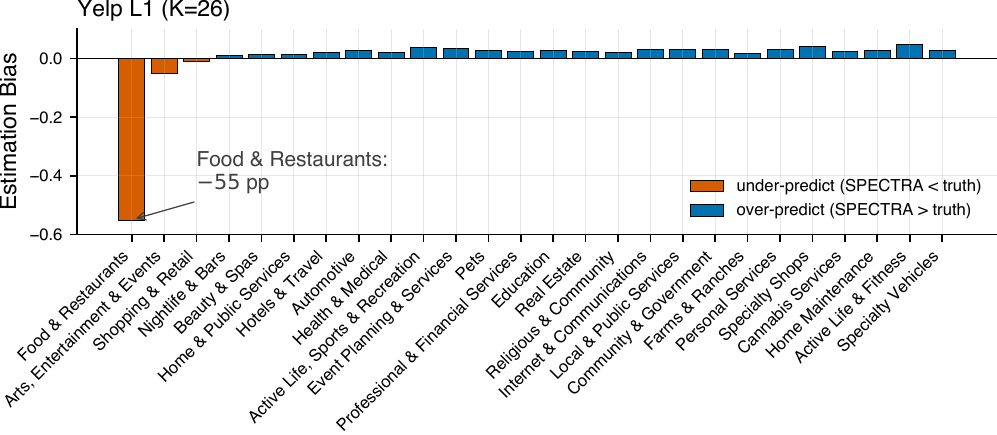}
  \caption{Per-category bias of SPECTRA, $\bar\theta_S -
  \bar\theta_*$, on Yelp L1 ($K{=}26$). Categories are sorted by true
  frequency $\bar\theta_*$ descending; orange bars indicate SPECTRA
  under-prediction, blue bars indicate over-prediction. The dominant
  Food \& Restaurants category ($63.9\%$ true mass) is under-predicted
  by $55.2\%$, dwarfing the other shifts. Companion
  MovieLens panel is \cref{fig:per_cluster_bias} in the main text.}
  \label{fig:per_cluster_bias_yelp_appendix}
\end{figure*}

\Cref{fig:per_cluster_bias_yelp_appendix} plots per-category bias for
Yelp L1; the MovieLens panel appears in the main text
(\cref{fig:per_cluster_bias}).

We expand the head/tail-stratified ECE into a per-category scatter of
mean predicted probability against mean empirical frequency, one point
per category, color-coded by head/mid/tail bucket. The systematic shift
of mass from heads (under-predicted by SPECTRA) toward tails
(over-predicted by SPECTRA) is visible as a coherent slope rather than
as scattered noise: this is the mass-redistribution mechanism in figure
form.

On MovieLens the most under-predicted categories are
Drama ($-7.9$\%; mean predicted $0.087$ vs.\ truth $0.166$),
Action ($-5.9$), Comedy ($-4.0$), Adventure ($-2.4$), and Sci-Fi
($-2.2$); the most over-predicted are
Film-Noir ($+3.3$\%; mean predicted $0.036$ vs.\ truth $0.002$),
Musical ($+3.1$), Mystery ($+2.7$), Documentary ($+2.5$), and
Animation ($+2.4$). The head/tail asymmetry is one-sided in sign --- no
head category is over-predicted, and no tail category is under-predicted
by more than $\sim 0.5$\%.

On Yelp L1 the dominant under-predicted category is Food \& Restaurants
($-55.2$\%; mean predicted $0.087$ vs.\ truth $0.639$); other
under-predicted heads are Arts, Entertainment \& Events ($-5.0$),
Shopping \& Retail ($-1.0$). The most over-predicted categories are
Active Life \& Fitness ($+4.8$\%; truth essentially zero),
Specialty Shops ($+4.2$), Active Life, Sports \& Recreation ($+3.8$),
Event Planning \& Services ($+3.5$), and Personal Services ($+3.3$).
The Yelp pattern matches MovieLens in direction but is amplified by
the Pareto-extreme nature of Yelp's empirical distribution.

\subsection{Aggregate vs.\ stratified ECE}
\label{sec:appendix:calibration:ece}

\Cref{tab:ece_stratified} promotes the aggregate-and-stratified
ECE numbers used in the main text. Aggregate ECE is $0.0043$ on MovieLens and $0.0286$ on Yelp L1.
Head-stratified ECE is larger on both datasets ($9.4\times$ aggregate
on MovieLens and $2.3\times$ on Yelp L1). The category-bias plots show
the direction of these errors.

\begin{table}[h]
\centering
\caption{Soft-label ECE, aggregate and stratified by head/mid/tail
buckets. Head/mid/tail buckets are defined by tertiles of the
aggregate empirical mass $\theta_*(c)$; counts of category-user pairs
$n$ are reported per stratum.}
\label{tab:ece_stratified}
\small
\begin{tabular}{lrrr}
\toprule
\textbf{Dataset / stratum} & \textbf{ECE} & \textbf{n pairs} & \textbf{categories} \\
\midrule
MovieLens, overall & $0.0043$ & $379{,}962$ & $19$ \\
MovieLens, head    & $0.0407$ & $119{,}988$ & $6$ \\
MovieLens, mid     & $0.0163$ & $119{,}988$ & $6$ \\
MovieLens, tail    & $0.0209$ & $139{,}986$ & $7$ \\
\midrule
Yelp L1, overall   & $0.0286$ & $259{,}896$ & $26$ \\
Yelp L1, head      & $0.0658$ & $79{,}968$  & $8$ \\
Yelp L1, mid       & $0.0275$ & $89{,}964$  & $9$ \\
Yelp L1, tail      & $0.0310$ & $89{,}964$  & $9$ \\
\bottomrule
\end{tabular}
\end{table}

\subsection{ECE by activity stratum (per-user)}
\label{sec:appendix:calibration:per_user}

A per-user breakdown of ECE by activity stratum (history length
quartile) is consistent with the head/tail picture: short-history users
show ECE patterns dominated by head under-prediction on their few
observed categories, while long-history users show a wider distribution of
ECE across the category vocabulary. We do not include a separate
activity-stratum-stratified ECE table here as the head/tail-stratified
results already characterize the asymmetry; the per-user
JS-divergence stratification by tail-preference quartile in
\cref{sec:experiments} is the user-level analog.

\section{Robustness ablations}
\label{sec:appendix:robustness}

This appendix presents five robustness ablations (R1--R5) addressing
taxonomy sensitivity, prompt and verbalizer sensitivity, calibration,
latency, and industrial distributional alignment.
\Cref{sec:appendix:robustness:taxonomy} answers R1 (taxonomy
sensitivity); \cref{sec:appendix:robustness:prompt} answers R2 (prompt and
verbalizer sensitivity); \cref{sec:appendix:robustness:calibration_diag}
answers R3 (top-$1$ ECE calibration diagnostic);
\cref{sec:appendix:robustness:latency} answers R4 (wall-clock latency);
\cref{sec:appendix:robustness:industrial} answers R5 (industrial
cohort-level JS divergence).

\subsection{R1 -- Taxonomy granularity robustness}
\label{sec:appendix:robustness:taxonomy}

We perturb the MovieLens taxonomy by semantically adjacent merges from
$K=19$ down to $K=15$ (merge Adventure$\to$Action, Fantasy$\to$Sci-Fi,
Mystery$\to$Crime, Horror$\to$Thriller) and then to $K=10$ (further
coarsening of pair-wise adjacent categories). The table reports the probe's long- and short-term results for a
five-session context window.

\begin{table}[h]
\centering
\caption{Taxonomy granularity robustness on MovieLens (Qwen3-8B,
CPT+SFT, CW$=5$). $K{=}19$ is the original taxonomy.
$K{=}15$ and $K{=}10$ merge semantically related categories. The
support changes with $K$, so the metrics describe different prediction tasks.}
\label{tab:r1_taxonomy}
\small
\resizebox{\columnwidth}{!}{%
\begin{tabular}{lccc}
\toprule
\textbf{Taxonomy} & \textbf{CW} & \textbf{LT NDCG@10 / JS} & \textbf{ST NDCG@10 / JS} \\
\midrule
$K{=}19$ (canonical) & 5 & $0.825$ / $0.315$ & $0.781$ / $0.460$ \\
$K{=}15$             & 5 & $0.87$ / $0.32$   & $0.84$ / $0.43$   \\
$K{=}10$             & 5 & $0.90$ / $0.25$   & $0.86$ / $0.36$   \\
\bottomrule
\end{tabular}}
\end{table}

\subsection{R2 -- Prompt and verbalizer sensitivity}
\label{sec:appendix:robustness:prompt}

We report two probe-template perturbations on MovieLens with the
canonical $K=19$ taxonomy: an $N{=}1$ prompt paraphrase of the
likelihood-probe template (rewording while preserving intent and
verbalizer set) and a simplified verbalizer ablation (the canonical
$V_+ = \{\texttt{Yes}, \texttt{Y}, \texttt{y}\}$ and
$V_- = \{\texttt{No}, \texttt{N}, \texttt{n}\}$ reduced to the
single-token affirmative/negative pair $\{\texttt{Yes}\}, \{\texttt{No}\}$).
NDCG@$10$ values across two context windows:

\begin{table}[h]
\centering
\caption{Prompt-paraphrase and simplified-verbalizer robustness on
MovieLens (NDCG@$10$). $N{=}1$ paraphrase and a single simplified
verbalizer set are reported; the canonical SPECTRA row reproduces the
main-text headline.}
\label{tab:r2_prompt}
\small
\resizebox{\columnwidth}{!}{%
\begin{tabular}{lcccc}
\toprule
\textbf{Setting} & \textbf{CW$=$3 LT} & \textbf{CW$=$3 ST} & \textbf{CW$=$5 LT} & \textbf{CW$=$5 ST} \\
\midrule
DG               & $0.79$ & $0.56$ & $0.76$ & $0.70$ \\
SPECTRA (canon.) & $0.86$ & $0.76$ & $0.86$ & $0.77$ \\
SPECTRA (paraphrase) & $0.86$ & $0.76$ & $0.86$ & $0.77$ \\
SPECTRA (simpl.\ verb.) & $0.85$ & $0.77$ & $0.86$ & $0.78$ \\
\bottomrule
\end{tabular}}
\end{table}

Both perturbations preserve the SPECTRA advantage with small
($\le 0.02$ NDCG@$10$) variance relative to the canonical prompt. These results cover one paraphrase and one simpler verbalizer set.

\subsection{R3 -- Top-$1$ ECE calibration diagnostic}
\label{sec:appendix:robustness:calibration_diag}

Beyond the head/tail-stratified ECE used in
\cref{sec:appendix:calibration:ece}, we report a top-$1$ ECE adapted for
multi-category supervision: for each user, compare the model's top-$1$
predicted probability to the empirical frequency of the corresponding
category, then bin and average.

\begin{table}[h]
\centering
\caption{Top-$1$ ECE for SPECTRA logit-probing on MovieLens, four
operating points.}
\label{tab:r3_top1_ece}
\small
\begin{tabular}{lc}
\toprule
\textbf{Setting} & \textbf{Top-1 ECE} \\
\midrule
CW$=$3, long-term  & $0.0036$ \\
CW$=$3, short-term & $0.0047$ \\
CW$=$5, long-term  & $0.0022$ \\
CW$=$5, short-term & $0.0022$ \\
\bottomrule
\end{tabular}
\end{table}

Top-$1$ ECE stays below $0.005$ in these four settings.
\Cref{sec:appendix:calibration} reports error by category group.

\subsection{R4 -- Wall-clock latency}
\label{sec:appendix:robustness:latency}

\begin{table}[h]
\centering
\caption{Wall-clock latency for DG and SPECTRA, measured on a single
A100-80G. The $K$-linear scaling of likelihood
probing is the cost of the distributional read-out; production
deployments amortize this in batched offline pipelines. The inference jobs reported elsewhere
in this submission use L40S, not A100; the relative cost of probing
versus DG is preserved across GPU families.}
\label{tab:r4_latency}
\small
\resizebox{\columnwidth}{!}{%
\begin{tabular}{lcc}
\toprule
\textbf{Method} & \textbf{Per-query latency} & \textbf{Notes} \\
\midrule
DG (uniform-on-top-$5$)               & $0.101$\,s   & single fwd + $k$-token decode \\
SPECTRA, no prefill cache             & $\approx 0.10 \cdot K$\,s & $K$ probing prompts \\
SPECTRA, with prefill cache           & $\approx 0.05 \cdot K$\,s & shared history prefix \\
\bottomrule
\end{tabular}}
\end{table}

Probing cost is amortized in batched offline pipelines: production
deployments compute SPECTRA distributions on a daily update cadence and
serve cached distributions at request time, so the per-request overhead
at serve time is zero. The complexity analysis in
\cref{sec:appendix:compute} supports
this picture with token-length proxy ratios at different
prefix/suffix regimes.

\subsection{R5 -- Industrial distributional alignment}
\label{sec:appendix:robustness:industrial}

\Cref{tab:r5_industrial} compares the two models' category distributions
at the cohort level. The short-video ranking comparison appears in
\cref{sec:exp_downstream}.

\begin{table}[h]
\centering
\caption{Cohort-level JS divergence ($\downarrow$) on the proprietary
short-video subset, using 30 days of history and a 14-day prediction window.}
\label{tab:r5_industrial}
\small
\begin{tabular}{lc}
\toprule
\textbf{Method} & \textbf{JS divergence} \\
\midrule
SPECTRA & $0.47$ \\
Production behavior-sequence ranker & $0.70$ \\
\bottomrule
\end{tabular}
\end{table}

SPECTRA's cohort-level JS divergence is $33\%$ lower. This metric
compares the full category distribution in aggregate.

\section{Downstream item-utility demo}
\label{sec:appendix:downstream_utility}

We test whether SPECTRA's category-level distribution can act as a
useful prior on top of classical sequential recommenders, by fusing
it with two backbones, SASRec~\citep{kang2018self} and
GRU4Rec~\citep{hidasi2015session}, on MovieLens. Both backbones are
trained on the non-evaluation MovieLens training split; SPECTRA scores come
from the same Qwen3-8B$+$CPT$+$SFT used elsewhere. Final scores fuse the
base ranker and the SPECTRA category affinity as $\text{final} =
z_\text{base} + \lambda \cdot z_\text{SPECTRA}$, sweeping $\lambda$
over $\{0, 0.05, 0.1, 0.5, 1, 2\}$. We evaluate on the user subset with a
non-empty test window, using two candidate-set sizes (1 true $+ 99$
negatives, and 1 true $+ 999$ negatives).

\Cref{tab:appendix:downstream} reports the fusion result at the best
weight $\lambda{=}0.5$. SPECTRA as a prior gives a small but
significant overall lift on both backbones, and a much larger lift
on the mid-popularity slice (items in the middle third by global
popularity), where the base ranker is weakest.

\begin{table}[h]
\centering
\caption{\textit{SPECTRA as a fused category prior on MovieLens.}
NDCG@$10$ with $\lambda{=}0$ (base ranker alone) versus $\lambda{=}0.5$
(best fusion weight). Paired-Wilcoxon $p$-values are within-user,
$n_{\text{paired}}=17{,}498$.}
\label{tab:appendix:downstream}
\small
\resizebox{\columnwidth}{!}{%
\begin{tabular}{lccc}
\toprule
\textit{Backbone / slice} & \textit{base alone} & \textit{$+$ SPECTRA} & $p$ \\
\midrule
SASRec, neg99        & $0.2472$ & $\mathbf{0.2483}$ & $<10^{-12}$ \\
SASRec, neg999       & $0.1253$ & $\mathbf{0.1265}$ & $<10^{-12}$ \\
GRU4Rec, neg99       & $0.2916$ & $\mathbf{0.3003}$ & $<10^{-23}$ \\
GRU4Rec, neg999      & $0.1466$ & $\mathbf{0.1474}$ & $\sim 0.04$ \\
\midrule
SASRec mid, neg99    & $0.0356$ & $\mathbf{0.0497}$ & $<10^{-12}$ \\
GRU4Rec mid, neg99   & $0.0369$ & $\mathbf{0.0596}$ & $<10^{-23}$ \\
\bottomrule
\end{tabular}}
\end{table}

The mid-slice lift ($+40\%$ relative on SASRec, $+62\%$ relative on
GRU4Rec) is the load-bearing finding. SPECTRA does not change the
head ranking, where the base ranker already has dense signal, but
boosts mid-popularity items the base ranker under-ranks. Whether
SPECTRA adds value on a given dataset therefore depends on how much
of the evaluation mass sits on mid- and tail-popularity items, which
is a property of the target distribution and not of SPECTRA itself.

\paragraph{Category-history priors.}
A per-user category-history frequency prior (often called UHF), along
with Laplace-smoothed and first-order Markov variants, is a useful
stationarity diagnostic for category-distribution prediction. Such
priors are constructed from each user's own past category counts and
therefore answer a different question from the one targeted by
SPECTRA: how well tomorrow's category mix can be replayed from
yesterday's, rather than whether an aligned LLM exposes a calibrated
preference distribution through its logits. On a small flat category
space such as MovieLens $K{=}19$, where histories are dense and
category distributions are relatively stable, these priors can be
strong sanity checks. We therefore treat history-count baselines as
complementary diagnostics for stationarity, and use DG,
Qwen3-Reranker, SASRec, and GRU4Rec as the load-bearing baselines for
the LLM-readout and downstream-utility claims evaluated here.

\section{Theory proofs}
\label{sec:appendix:proofs}

This appendix proves the main ranking result in
\cref{thm:ranking-advantage} and gives a complementary comparison in
total variation.

\textbf{Notation.} We use $\theta_*$ and $\theta_S$ as defined in
\cref{sec:method:setup,sec:method:primitive}. Let $\theta_{\mathrm{DG}}$
be the distribution supported on the decoded list $L \subseteq C$,
where $|L|=k<K$. Total variation is
$\|p-q\|_{\mathrm{TV}} := \frac{1}{2}\sum_c |p(c)-q(c)|$.
Write $m_\tau(L) := \sum_{c \notin L} \theta_*(c)$ for the true mass
outside the list.

\subsection{Lemma 1 -- DG truncation TV lower bound (proof)}
\label{sec:appendix:proofs:lemma1}

\begin{lemma}[DG truncation TV lower bound; tight, decoder-agnostic]
\label{lem:dg-truncation}
For any DG-induced distribution $\theta_{\mathrm{DG}}$ supported on a
decoded list $L \subseteq C$ with $|L|=k<K$ and any underlying allocation
within $L$,
\[
\|\theta_{\mathrm{DG}} - \theta_*\|_{\mathrm{TV}} \;\geq\;
m_\tau(L).
\]
\end{lemma}

\begin{proof}[Proof (TV-sup, headline tight form).]
Using the supremum representation
$\|p-q\|_{\mathrm{TV}} = \sup_{A \subseteq C} |p(A)-q(A)|$
(an equivalence between $\ell_1$ and event-supremum forms of TV),
take the test event $A := L$. Since
$\mathrm{supp}(\theta_{\mathrm{DG}}) \subseteq L$,
$\theta_{\mathrm{DG}}(L) = 1$; by the definition of $m_\tau$,
$\theta_*(L) = 1 - m_\tau(L)$. Hence
\[
\|\theta_{\mathrm{DG}} - \theta_*\|_{\mathrm{TV}} \;\geq\;
|\theta_{\mathrm{DG}}(L) - \theta_*(L)| \;=\; m_\tau(L). \qedhere
\]
\end{proof}

\textbf{Equivalent $\ell_1$-balance derivation (slack form).}
An alternate route through the $\ell_1$ representation of TV gives
the slack form $\|\theta_{\mathrm{DG}} - \theta_*\|_{\mathrm{TV}} \ge
m_\tau(L)/2$: starting from
$2\|\theta_{\mathrm{DG}} - \theta_*\|_{\mathrm{TV}} =
\sum_c|\theta_{\mathrm{DG}}(c) - \theta_*(c)| \ge
\sum_{c \notin L} \theta_*(c) = m_\tau(L)$
and discarding the in-list summand $\sum_{c \in L}|\theta_{\mathrm{DG}}(c)
- \theta_*(c)|$. This form is admissible but strictly looser; we use
the tight TV-sup form throughout.

\textbf{Equality conditions.}
Equality in the tight bound holds iff $\theta_{\mathrm{DG}}(c) \geq
\theta_*(c)$ for every $c \in L$, characterizing the $(k{-}1)$-simplex
$E(L; \theta_*)$ of equality-attaining distributions. For $m_\tau(L)<1$, renormalizing $\theta_*$ on $L$ attains equality.
For $m_\tau(L)=1$, every distribution supported on $L$ attains equality.

\textbf{Remarks.} The bound is \emph{decoder-agnostic}: it holds for greedy, top-$k$,
nucleus, beam, and uniform-on-top-$k$ DG decoders alike. The
operational empirical contrast in our main experiments uses
uniform-on-top-$5$ (\cref{sec:appendix:repro:dg}). The same identity
for the renormalized distribution
appears as Lemma~4.1 of \citet{tzachristas2025topk}, independently
derived in the different setting of sparse-attention truncation inside
transformer blocks.

\subsection{Lemma 2 -- SPECTRA Pinsker upper bound (proof)}
\label{sec:appendix:proofs:lemma2}

\begin{lemma}[SPECTRA Pinsker calibration upper bound]
\label{lem:spectra-calibration}
If $\mathrm{KL}(\theta_* \,\|\, \theta_S) \leq \delta_{\mathrm{KL}}$
with KL measured in nats, then
\[
\|\theta_S - \theta_*\|_{\mathrm{TV}} \;\leq\;
\sqrt{\delta_{\mathrm{KL}}/2}.
\]
\end{lemma}

\begin{proof}
This is Pinsker's inequality
\citep{pinsker1960information,canonne2022short} applied to the pair
$(\theta_*, \theta_S)$.
\end{proof}

To bound TV using JS directly, define $a=(p+q)/2$. Then
$\|p-a\|_{\mathrm{TV}}=\|q-a\|_{\mathrm{TV}}=\|p-q\|_{\mathrm{TV}}/2$.
Pinsker's inequality applied to $(p,a)$ and $(q,a)$ gives
\[
\|p-q\|_{\mathrm{TV}} \leq
\sqrt{2\ln(2)\,\mathrm{JS}_{\mathrm{bits}}(p,q)}.
\]
The factor $\ln(2)$ converts JS from bits to nats.

\subsection{Theorem -- composition of Lemma~1 and Lemma~2}
\label{sec:appendix:proofs:theorem}

\begin{theorem}[Regime-conditional dominance in total variation]
\label{thm:regime-dominance}
Suppose
\begin{itemize}
  \item[(A1)] SPECTRA's calibration error is bounded:
        $\delta_{\mathrm{JS}} := \mathrm{JS}_{\mathrm{bits}}(\theta_*, \theta_S)
        \leq \tilde\delta$ for some operational upper bound
        $\tilde\delta$.
  \item[(A2)] The true preference distribution has tail mass beyond
        DG's decoded set: $m_\tau(L) \geq \mu$ for some $\mu > 0$.
\end{itemize}
If, in addition, the \emph{regime condition} holds:
\[
\mu \;>\; \sqrt{2\ln(2)\,\tilde\delta},
\]
then SPECTRA's TV distance to $\theta_*$ is strictly less than DG's:
\[
\|\theta_S - \theta_*\|_{\mathrm{TV}}
\;<\;
\|\theta_{\mathrm{DG}} - \theta_*\|_{\mathrm{TV}}.
\]
\end{theorem}

\begin{proof}
By \cref{lem:dg-truncation}, $\|\theta_{\mathrm{DG}}-\theta_*\|_{\mathrm{TV}}
\geq m_\tau(L) \geq \mu$. The JS bound in
\cref{sec:appendix:proofs:lemma2} gives
$\|\theta_S-\theta_*\|_{\mathrm{TV}} \leq \sqrt{2\ln(2)\,\tilde\delta}
< \mu$, proving the claim.
\end{proof}

\subsection{Ranking-advantage theorem (proof and Schur-isotone generalization)}
\label{sec:appendix:ranking-advantage}

\Cref{thm:ranking-advantage} in \cref{sec:method:theory} gives the
headline per-user ranking statement. We collect here the proof in
slightly fuller form, the extension to a broader family of ranking
metrics, and a discussion of failure modes.

\textbf{Proof of \cref{thm:ranking-advantage} (detailed).}
Fix a user satisfying (i)--(iii). For discrete distributions on $C$,
$\|\theta_S - \theta_*\|_\infty = \max_c |\theta_S(c) - \theta_*(c)|
\leq \|\theta_S - \theta_*\|_{\mathrm{TV}}$, so any TV upper bound
gives a valid $\varepsilon$ certificate; this is how
\cref{lem:spectra-calibration} feeds into (i). For
$c \in G^*$ and $c' \in C \setminus G^*$,
\begin{align*}
\theta_S(c) - \theta_S(c')
&\geq \big(\theta_*(c) - \varepsilon\big) - \big(\theta_*(c') + \varepsilon\big) \\
&\geq \theta_*^{(k)} - \theta_*^{(k+1)} - 2\varepsilon \\
&= \Delta_k - 2\varepsilon \;>\; 0
\end{align*}
by (ii), proving that every $c \in G^*$ strictly outranks every
$c' \notin G^*$ under $\theta_S$. Hence $L_S = G^*$. The mass
inequality follows from a swap argument: writing
$j := |G^* \setminus L| = |L \setminus G^*| \geq 1$,
$M(G^*, \theta_*) - M(L, \theta_*) = \sum_{c \in G^* \setminus L} \theta_*(c)
- \sum_{c \in L \setminus G^*} \theta_*(c) \geq j \cdot \Delta_k > 0$. $\square$

\subsubsection{Schur-isotone generalization}
\label{sec:appendix:ranking-advantage:schur}

The argument extends to a family of ranking-quality metrics beyond
mass-captured.

\begin{definition}[Schur-isotone metric]
A ranking-quality metric $F(\pi, \theta_*)$ on size-$k$ lists $\pi$ is
\emph{Schur-isotone} if it takes one of the forms:
(\emph{Type-S, set-based}) $F(\pi, \theta_*) = \sum_{c \in \pi} f(\theta_*(c))$
for a monotone non-decreasing $f$;
(\emph{Type-O, order-aware}) $F(\pi, \theta_*) = \sum_{i=1}^k w_i \,f(\theta_*(\pi_i))$
for monotone non-decreasing $f$ and non-increasing weights
$w_1 \geq \dots \geq w_k \geq 0$.
\end{definition}

Mass-captured $M$ is Type-S with $f(x) = x$. Type-O instances include
DCG@$k$ ($f(x) = x$, $w_i = 1/\log_2(i+1)$) and DCG with exponential
gain ($f(x) = 2^x - 1$). NDCG@$k$ equals DCG@$k$ divided by an ideal
DCG which depends on $\theta_*$ alone, so Schur-isotonicity transfers.

\begin{theorem}[Generalized ranking advantage]
\label{thm:ranking-advantage-schur}
Let $F$ be a Schur-isotone ranking metric with strictly increasing
gain $f$ and, for Type-O, positive rank weights. Under (i)--(iii) of
\cref{thm:ranking-advantage} with $\varepsilon$ replaced (for Type-O
$F$) by $\varepsilon < \tfrac{1}{2}\min\!\big(\Delta_k,\,\Delta_k^{\min}\big)$
where $\Delta_k^{\min} := \min_{1 \leq i < j \leq k}(\theta_*^{(i)} - \theta_*^{(j)})$
is the minimum within-top-$k$ pairwise gap, we have
$F(L_S, \theta_*) > F(L, \theta_*)$.
\end{theorem}

\begin{proof}[Proof sketch]
For Type-S $F$ with monotone $f$ and $\theta_S$ recovering $G^*$ as a
set (Step 1 of \cref{thm:ranking-advantage}'s proof), the rearrangement
inequality gives $F(G^*, \theta_*) \geq F(L, \theta_*)$ with strict
inequality follows from strict increase of $f$ and $L \neq G^*$.
For Type-O $F$, the
stronger gap condition ensures within-$G^*$ argmax stability (applying
the same calibration argument to each pairwise gap inside $G^*$), so
$L_S$ recovers $G^*$ in order. Replacing omitted top-$k$ categories
strictly increases the score because the gains are strictly increasing
and the weights are positive. Sorting by decreasing gain can only
increase it further.
\end{proof}

\textbf{Caveat.} The stronger gap condition $\Delta_k^{\min} > 2\varepsilon$
is more restrictive than $\Delta_k > 2\varepsilon$. Type-O Schur-isotonicity
is therefore a property of distributions with well-separated top-$k$
ranks, not just well-separated top-$k$ sets.

\subsubsection{Population-expectation and finite-sample forms}
\label{sec:appendix:ranking-advantage:population}

The per-user deterministic statement
(\cref{thm:ranking-advantage}) admits two probabilistic lifts.

\textbf{Expectation form.}
Let $\mathcal{F} := \{u : \text{(i)--(iii) hold}\}$ be the favorable
cohort, and let
$\mathrm{Gain}_F(u) := F(L_S^u, \theta_*^u) - F(L^u, \theta_*^u)$.
With $F_{\max} := \sup_\pi F(\pi, \cdot) - \inf_\pi F(\pi, \cdot)$,
splitting the expectation over $\mathcal{F}$ and its complement gives
\begin{equation}
\begin{split}
\mathbb{E}[\mathrm{Gain}_F(u)] &\geq \Pr[\mathcal{F}] \cdot \mathbb{E}[\mathrm{Gain}_F \mid \mathcal{F}] \\
&\quad - F_{\max} \cdot (1 - \Pr[\mathcal{F}]).
\end{split}
\label{eq:ranking-expectation}
\end{equation}
The right-hand side is strictly positive whenever
$\Pr[\mathcal{F}] > F_{\max} / (F_{\max} + \mathbb{E}[\mathrm{Gain}_F \mid \mathcal{F}])$.

\textbf{Finite-sample CI.}
For $n$ i.i.d. users with $|\mathrm{Gain}_F| \leq F_{\max}$, Hoeffding's
inequality gives
\begin{equation}
\big|\widehat{\mathrm{Gain}}_F^{(n)} - \mathbb{E}[\mathrm{Gain}_F]\big|
\;\leq\; F_{\max}\sqrt{2\log(2/\alpha)/n}
\label{eq:ranking-hoeffding}
\end{equation}
with probability at least $1 - \alpha$. Massart's tight DKW yields the
analogous CI for $\widehat{\Pr}_n[\mathcal{F}]$ with $F_{\max} = 1$.

\subsubsection{Calibration certificate family}
\label{sec:appendix:ranking-calibration-family}

\Cref{thm:ranking-advantage} requires any pointwise upper bound
$\|\theta_S - \theta_*\|_\infty \leq \varepsilon$. Tighter
certificates yield a larger favorable cohort; we list admissible
choices below.

\begin{itemize}
\item \emph{Direct $\ell_\infty$ measurement} (tightest, requires
per-category access to $\theta_*$).
\item \emph{Total variation:} $\|\theta_S - \theta_*\|_\infty \leq \|\theta_S - \theta_*\|_{\mathrm{TV}}$.
\item \emph{JS:}
$\varepsilon = \sqrt{2\ln(2)\,\mathrm{JS}_{\mathrm{bits}}(\theta_S,\theta_*)}$
by the midpoint argument in \cref{sec:appendix:proofs:lemma2}.
\item \emph{Pinsker on KL:} $\varepsilon = \sqrt{\mathrm{KL}(\theta_* \,\|\, \theta_S)/2}$.
\item \emph{Hellinger:} $\varepsilon = \sqrt{2}\,H(\theta_S, \theta_*)$.
\item \emph{$\chi^2$:} $\varepsilon = \sqrt{\chi^2(\theta_*, \theta_S)/2}$.
\end{itemize}
The theorem statement is divergence-agnostic: any certificate above
suffices in (i), with the trade-off that looser certificates yield
smaller favorable cohorts.

\subsubsection{Failure modes}
\label{sec:appendix:ranking-advantage:failure}

The favorable cohort $\mathcal{F}$ excludes users with any of:
\begin{itemize}
\item \emph{Calibration failure}: $\varepsilon \geq \Delta_k / 2$.
\item \emph{Small ranking gap}: $\Delta_k$ near zero (near-tie between
the $k$-th and $(k{+}1)$-th categories). The theorem is silent on
such users; their ranking under $\theta_S$ may or may not equal
$G^*$.
\item \emph{DG-recovers-truth}: $L = G^*$. SPECTRA still recovers
$G^*$, but the strict inequality on $M$ degenerates to equality.
\end{itemize}
Reporting per-dataset $\Pr[\mathcal{F}]$ and conditional gain
$\mathbb{E}[\mathrm{Gain} \mid \mathcal{F}]$ separates these effects.

\section{Stage-wise probing analysis}
\label{sec:appendix:stagewise}

This appendix runs the SPECTRA probe at three checkpoints of the LM
training pipeline to locate where category-level preference structure
emerges: (i) the pretrained backbone, (ii) the SFT-only adapter
(trained directly on top of the pretrained backbone without continued
pretraining), and (iii) the full CPT$+$SFT adapter used in the main
experiments. The CPT$+$SFT stage corresponds to the probe results
already reported in \cref{sec:experiments} and \cref{sec:appendix:calibration};
this appendix is a stage-decomposition of the same probe.

\subsection{Methodology}
\label{sec:appendix:stagewise:method}

The MovieLens comparison uses the same probing prompts and verbalizer
sets at each stage. The pretrained backbone has no adapter. The two
adapted stages use the LoRA configuration described in
\cref{sec:appendix:repro:checklist} (C8).

\subsection{Stage-wise results on MovieLens}
\label{sec:appendix:stagewise:results}

\Cref{tab:stagewise_ablation} reports SPECTRA Logit-Probing
performance across the three training stages on MovieLens
($n{=}20{,}000$ probe records per stage, Qwen3-8B backbone, anchor
context window $\mathrm{CW}{=}8$). The pretrained-only backbone's
mean predicted category masses span a range of $0.030$.

\begin{table}[h]
\centering
\caption{\textit{Stage-wise ablation on MovieLens} (Qwen3-8B, CW=8).
NDCG@10 ($\uparrow$) and Jensen-Shannon divergence ($\downarrow$)
for the SPECTRA Logit-Probing readout under three training pipelines.
Long-term (LT) and short-term (ST) prediction horizons.}
\label{tab:stagewise_ablation}
\small
\resizebox{\columnwidth}{!}{%
\begin{tabular}{l cc cc}
\toprule
& \multicolumn{2}{c}{\textit{Long-term}}
& \multicolumn{2}{c}{\textit{Short-term}} \\
\cmidrule(lr){2-3}\cmidrule(lr){4-5}
\textit{Training stage} & NDCG@10 & JS-Div & NDCG@10 & JS-Div \\
\midrule
Pretrain-only             & $0.889$ & $0.338$ & $0.727$ & $0.498$ \\
SFT-only (no CPT)         & $\mathbf{0.893}$ & $0.366$ & $\mathbf{0.807}$ & $0.499$ \\
CPT$+$SFT (main paper)    & $0.863$ & $\mathbf{0.316}$ & $0.780$ & $\mathbf{0.462}$ \\
\bottomrule
\end{tabular}}
\end{table}

\paragraph{Reading.} The full CPT$+$SFT pipeline used in the main
paper achieves the best \emph{distributional alignment}
(lowest JS divergence on both LT and ST), confirming that domain
continued pretraining contributes meaningfully to calibration
beyond SFT alone. SFT-only matches or slightly exceeds CPT$+$SFT
on top-$k$ ranking accuracy (NDCG@10), suggesting that the ranking
signal is largely carried by the supervised fine-tuning stage,
while CPT contributes the distributional fidelity that drives the
main paper's alignment and long-tail-recovery results. The
pretrained-only backbone is competitive on long-term NDCG but
substantially weaker on short-term prediction, reflecting the
absence of any task-specific adaptation.

\section{Computational complexity}
\label{sec:appendix:compute}

We retain the complexity analysis from prior work as it supports
R4 latency (\cref{sec:appendix:robustness:latency}). The summary is:
likelihood probing scales as
$O(C(\mathcal{M},T_{\mathrm{pre}}) + K\cdot C(\mathcal{M},
T_{\mathrm{suf}}))$ with KV-cache prefix reuse;
generative classification scales as $O(C(\mathcal{M},T))$ (single-shot);
hierarchical probing scales as
$O(K_1\cdot C(\mathcal{M}, T_{L1}) + B\cdot K_{2,\mathrm{avg}}\cdot
C(\mathcal{M},T_{L2}))$. Token-length
proxy ratios at typical regimes ($T_{\mathrm{pre}}=80$, $T_{\mathrm{suf}}=20$,
$K=20$) yield $\sim 4.8\times$ overhead under short user histories and
$\sim 1.9\times$ overhead under long user histories
($T_{\mathrm{pre}}=300$), measured as token-length ratios rather than
wall-clock seconds.

\end{document}